\documentclass[fleqn,10pt]{wlscirep}
\usepackage[utf8]{inputenc}
\usepackage[T1]{fontenc}

\usepackage[
sorting=none,
maxbibnames=99,
backend=biber
]{biblatex}
\BiblatexSplitbibDefernumbersWarningOff
\usepackage{csquotes}

\definecolor{citecolor}{HTML}{0071bc}
\definecolor{citered}{HTML}{8b0000}
\usepackage{hyperref}
\usepackage{url}

\usepackage{booktabs}

\usepackage{amsmath,amsfonts,bm}

\def\eqref#1{equation~\ref{#1}}

\def\1{\bm{1}}

\def\va{{\bm{a}}}

\def\ve{{\bm{e}}}

\def\vg{{\bm{g}}}
\def\vh{{\bm{h}}}

\def\vr{{\bm{r}}}

\def\vt{{\bm{t}}}
\def\vu{{\bm{u}}}
\def\vv{{\bm{v}}}

\def\vx{{\bm{x}}}
\def\vy{{\bm{y}}}
\def\vz{{\bm{z}}}

\DeclareMathAlphabet{\mathsfit}{\encodingdefault}{\sfdefault}{m}{sl}
\SetMathAlphabet{\mathsfit}{bold}{\encodingdefault}{\sfdefault}{bx}{n}

\usepackage{graphicx}
\usepackage{float}
\usepackage{amsfonts,amsmath,amssymb,amsthm}
\usepackage{cleveref}
\usepackage{multicol, multirow}
\usepackage{makecell}
\usepackage{caption}
\usepackage{subcaption}
\usepackage{adjustbox}
\usepackage{enumitem}
\usepackage{wrapfig}
\usepackage{algorithm}
\usepackage{algorithmic}
\usepackage{longtable}

\usepackage{bbm}
\usepackage{accents}

\usepackage{cleveref}

\usepackage{xcolor}
\definecolor{ForestGreen}{RGB}{34,139,34}
\newcommand{\BindingNet}[1]{BindingNet}
\newcommand{\MDModel}[1]{NeuralMD}

\DeclareUnicodeCharacter{0301}{*************************************}

\newcommand{\ie}{\em{i.e.}}
\newcommand{\eg}{\em{e.g.}}
\usepackage{xcolor}
\definecolor{ForestGreen}{RGB}{34,139,34}

\newcommand{\norm}[1]{\left\lVert#1\right\rVert}

\title{
A Multi-Grained Symmetric Differential Equation Model for Learning Protein-Ligand Binding Dynamics
}
\author[1]{Shengchao Liu$^*$}
\author[2]{Weitao Du$^*$}
\author[3]{Hannan Xu}
\author[4]{Yanjing Li}
\author[5]{Zhuoxinran Li}
\author[4]{Vignesh Bhethanabotla}
\author[4]{Divin Yan}
\author[1]{Christian Borgs$^\dagger$}
\author[4]{Anima Anandkumar$^\dagger$}
\author[6]{Hongyu Guo$^\dagger$}
\author[1]{Jennifer Chayes$^\dagger$}
\affil[1]{University of California Berkeley, Berkeley, CA 94720, United States}
\affil[2]{DAMO Academy, Bellevue, WA 98004, United States}
\affil[3]{University of Oxford, Oxford OX1 2JD, UK}
\affil[4]{California Institute of Technology, Pasadena, CA 91125, United States}
\affil[5]{University of Toronto, Toronto, ON M5S 1A1, Canada}
\affil[6]{National Research Council Canada, Ottawa, ON K1N 6N5, Canada}

\footnotetext{Equal contribution.}
\footnotetext{Joint supervision.}

\begin{abstract}
In drug discovery, molecular dynamics (MD) simulation for protein-ligand binding provides a powerful tool for predicting binding affinities, estimating transport properties, and exploring pocket sites. There has been a long history of improving the efficiency of MD simulations through better numerical methods and, more recently, by utilizing machine learning (ML) methods. Yet, challenges remain, such as accurate modeling of extended-timescale simulations. To address this issue, we propose NeuralMD, the first ML surrogate that can facilitate numerical MD and provide accurate simulations in protein-ligand binding dynamics. We propose a principled approach that incorporates a novel physics-informed multi-grained group symmetric framework. Specifically, we propose (1) the BindingNet model that satisfies group symmetry using vector frames and captures the multi-level protein-ligand interactions, and (2) an augmented neural differential equation solver that learns the trajectory under Newtonian mechanics. For the experiment, we design ten single-trajectory and three multi-trajectory binding simulation tasks. We demonstrate the efficiency and effectiveness of NeuralMD, achieving over 1K$\times$ speedup compared to standard numerical MD simulations. NeuralMD also outperforms all other ML approaches, achieving up to 15$\times$ reduction in reconstruction error and 70\% increase in validity. Additionally, we qualitatively illustrate that the oscillations in the predicted trajectories align more closely with ground-truth dynamics than those of other machine-learning methods. We believe NeuralMD paves the foundation for a new research paradigm in simulating protein-ligand dynamics.
\end{abstract}

\begin{document}
\flushbottom
\maketitle
\thispagestyle{empty}

\section{Introduction}
The molecular dynamics (MD) simulation for protein-ligand binding is one of the fundamental tasks in drug discovery~\cite{kairys2019binding,lahey2020simulating,yang2020predicting,volkov2022frustration}. Such simulations of binding systems are a key component of the drug discovery pipeline to select, refine, and tailor the chemical structures of potential drugs to enhance their efficacy and specificity. To simulate the protein-ligand dynamics, \textit{numerical MD} methods have been extensively developed~\cite{lifson1968consistent,kohn1999nobel}. However, the numerical MD methods are computationally expensive due to the expensive force calculations on individual atoms in a large protein-ligand system.

To alleviate this issue, \textit{machine learning (ML)} surrogates have been proposed to either augment or replace numerical MD methods to estimate the MD trajectories. However, all prior ML approaches for MD simulation are limited to single-system and not protein-ligand complex~\cite{behler2007generalized,chmiela2017machine,zhang2018deep,liu2024CrystalFlow}. A primary reason is the lack of large-scale datasets for protein-ligand binding (the first large-scale dataset with binding dynamics was released in May 2023~\cite{siebenmorgen2024misato}). Further, prior ML-based MD approaches limit to fitting the energy surface so as to study the MD dynamics on a small timestep ({\eg}, 1e-15 seconds)~\cite{schutt2018schnet,thomas2018tensor,klicpera2020fast,qiao2020orbnet,musaelian2022learning,liu2023symmetry,liu2023group,du2023molecule}, while simulation on a larger timestep ({\eg}, 1e-9 seconds) is needed for specific tasks, such as detecting the transient and cryptic states in binding dynamics~\cite{vajda2018cryptic}. However, current longer-time ML MD simulations are challenging due to the catastrophic buildup of errors over longer rollouts~\cite{fu2022forces}.

\begin{figure}[tb!]
\centering
\includegraphics[width=0.99\textwidth]{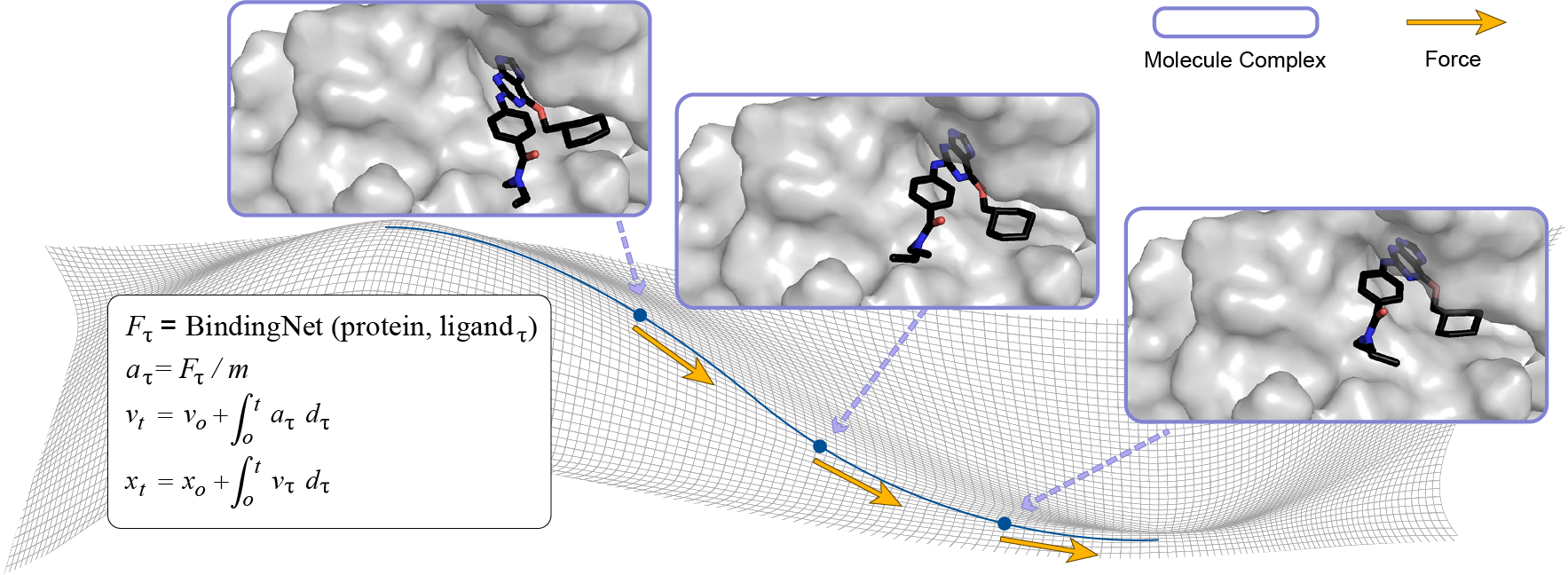}
\vspace{-2ex}
\caption{
Illustrations of \MDModel{} for binding dynamics. The landscape depicts the energy level, and the binding dynamic leads to an equilibrium state with lower energy. 
}
\label{fig:binding_dynamics_and_examples}
\vspace{-2ex}
\end{figure}

Another critical aspect that needs to be considered in ML-based modeling is the group symmetry present in the protein-ligand geometry. Specifically, the geometric function over molecular systems should be equivariant to rotation and translation, {\ie}, SE(3)-equivariance. One principled approach to satisfy equivariance is to use vector frames, which have been previously explored for single molecules~\cite{jumper2021highly}, but not yet for the binding complexes. The vector frame basis achieves SE(3)-equivariance by projecting vectors ({\eg}, positions and accelerations) to the vector frame basis, and such a projection can maintain the equivariant property with efficient calculations~\cite{liu2023symmetry}.

\textbf{Our Approach: \MDModel{}.} We propose \MDModel{}, a multi-grained physics-informed approach designed to handle extended-timescale MD simulations in protein-ligand binding. Our multi-grained method explicitly decomposes the large molecular complexes into three granularities to obtain an efficient approach for modeling the large molecular system: the atoms in ligands, the backbone structures in proteins, and the residue-atom pairs in binding complexes. We achieve group symmetry in \BindingNet{} through the incorporation of vector frames, and include three levels of vector frame bases for multi-grained modeling, from the atom and backbone level to the residue level for binding interactions.

Additionally, our ML approach, \MDModel{}, leverages data-driven techniques to learn Newtonian mechanics. In MD, the movement of atoms is determined by Newton's second law, $F = m \cdot a$, where $F$ is the force, $m$ is the mass, and $a$ is the acceleration of each atom. By integrating acceleration and velocity w.r.t. time, we can obtain the velocities and positions, respectively. Thus in \MDModel{}, we formulate the trajectory simulation as a second-order ordinary differential equation (ODE) or second-order stochastic differential equation (SDE) problem. We augment derivative space by concurrently calculating the accelerations and velocities, allowing simultaneous integration of velocities and positions.

\textbf{Experiments and Results.} To evaluate the effectiveness and efficiency of \MDModel{}, we design ten single-trajectory and three multi-trajectory binding simulation tasks. For quantitative assessment, we employed two reconstruction metrics and two validity metrics. The results demonstrate that \MDModel{} consistently outperforms other ML methods, achieving up to 15$\times$ reduction in reconstruction error and 70\% increase in validity. Additionally for qualitative analysis, we introduced a fluctuation measure to characterize ligand oscillation, supported by a video demonstration provided in the supplementary materials (also available at \href{https://chao1224.github.io/NeuralMD}{this link}). These results highlight that while \MDModel{} is not perfect, it generalizes more effectively than other ML baselines. Finally, in terms of efficiency, \MDModel{} delivers up to over 1K$\times$ speedup compared to numerical methods.

\section{Results}
\subsection{Preliminaries}
\textbf{Ligand data structure.}
In this work, we consider binding complexes involving small molecules as ligands. Small molecules can be treated as sets of atoms in the 3D Euclidean space, $\{f^{(l)}, \vx^{(l)}\}$, where $f^{(l)}$ and $\vx^{(l)}$ represent the atomic numbers and 3D Euclidean coordinates for atoms in each ligand, respectively.

\textbf{Protein data structure.}
Proteins are macromolecules, which are essentially chains of amino acids or residues. There are 20 natural amino acids, and each amino acid is a small molecule. Noticeably, amino acids are made up of three components: a basic amino group (-NH$_2$), an acidic carboxyl group (-COOH), and an organic R group (or side chain) that is unique to each amino acid. Additionally, the carbon that connects all three groups is called $\text{C}_\alpha$. Due to the large number of atoms in proteins, this work proposes a multi-grained method for modeling the protein-ligand complexes. In this regard, the \textbf{backbone-level} data structure for each protein is $\{f^{(p)}, \{\vx^{(p)}_{N}, \vx^{(p)}_{C_\alpha}, \vx^{(p)}_{C}\}\}$, for the residue type and the coordinates of $N-C_\alpha-C$ in each residue, respectively. We may omit the superscript in the coordinates of backbone atoms, as these backbone structures are unique to protein residues. In addition to the backbone level, as a coarser-grained view, we further consider \textbf{residue-level} information for modeling binding interactions, $\{f^{(p)}, \vx^{(p)}\}$, where the coordinate of $C_\alpha$ is taken as the residue-level coordinate, {\ie}, $\vx^{(p)} \triangleq \vx^{(p)}_{C_\alpha}$.

\textbf{Molecular Dynamics Simulations.}
Generally, molecular dynamics (MD) describes how each atom in a molecular system moves over time, following Newton's second law of motion:
\begin{equation}
\begin{aligned}
F = m \cdot \va = m \cdot \frac{d^2 \vx}{d t^2},
\end{aligned}
\end{equation}
where $F$ is the force, $m$ is the mass, $a$ is the acceleration, $\vx$ is the position, and $t$ is the time. Then, an MD simulation will take a second-order integration to get the trajectories for molecular systems like a small molecule, a protein, a polymer, or a protein-ligand complex. The \textbf{numerical MD} methods can be classified into \textbf{classical MD} and \textbf{\textit{ab-initio} MD}, where the difference lies in how the force on each atom is calculated: classical MD uses force field approaches to predict the atomic forces~\cite{lifson1968consistent}, while \textit{ab-initio} MD calculates the forces using quantum mechanical methods, such as density functional theory (DFT)~\cite{kohn1999nobel}. More recently, \textbf{ML MD} methods have opened a new perspective by utilizing the group symmetric tools for geometric representation and energy prediction~\cite{schutt2018schnet,thomas2018tensor,klicpera2020fast,qiao2020orbnet,musaelian2022learning,liu2023symmetry,liu2023group,du2023molecule,zhang2023artificial}, as well as the automatic differential tools for trajectory learning~\cite{chen2018neural,raissi2019physics,lu2019deeponet,lu2021learning,li2020neural,kovachki2021neural}. Please check Supplementary A for a more detailed discussion.

\textbf{Newtonian Dynamics and Langevin Dynamics.}
Numerical MD methods can be additionally divided into two categories: using Newtonian dynamics or Langevin dynamics. Newtonian dynamics is suitable for idealized systems with negligible thermal effects or when deterministic trajectories are required, while Langevin dynamics is adopted where thermal effects play a significant role and when the system is being studied at a finite temperature. In the ML-based MD simulations, adopting Newtonian dynamics or Langevin dynamics can be treated as an option for introducing different inductive biases. In this work, we propose two versions: an ordinary differential equation (ODE) solver and a stochastic differential equation (SDE) solver concerning Newtonian dynamics and Langevin dynamics, respectively. Noticeably, our experiential dataset, MISATO~\cite{siebenmorgen2024misato}, uses Newtonian dynamics with Langevin thermostats, and the information on solvent molecules is not provided.\looseness=-1

\textbf{Problem Setting: MD Simulation in Protein-Ligand Binding.}
In this work, we are interested in learning the MD simulation in the protein-ligand binding system, and we consider the semi-flexible setting~\cite{salmaso2018bridging}, {\ie}, proteins with rigid structures and ligands with flexible movements. Thus, the problem is formulated as follows: suppose we have a fixed protein structure $\{f^{(p)}, \{\vx^{(p)}_{N}, \vx^{(p)}_{C_\alpha}, \vx^{(p)}_{C}\}\}$ and a ligand with its initial structure and velocity, $\{f^{(l)}, \vx^{(l)}_0, \vv^{(l)}_0\}$. We want to predict the trajectories of ligands following the Newtonian dynamics, {\ie}, the movement of $\{\vx^{(l)}_{t}, ... \}$ over time. We also want to clarify two critical points about this problem setting: (1) Our task is trajectory prediction, {\ie}, positions as labels, and no explicit energy and force are considered as labels. ML methods for energy prediction followed with numerical ODE/SDE solver require smaller timestep ({\eg}, 1e-15 seconds), while trajectory prediction, which directly predicts the coordinates over time, is agnostic to the magnitude of the timestep. This is appealing for tasks with larger timestep ({\eg}, 1e-9 seconds), as will be discussed below. (2) Each trajectory is composed of a series of geometries of molecules, and such geometries are called \textbf{snapshots}. We avoid using \textit{frames} since we will introduce the \textit{vector frame} in modeling the binding complex in the following sections.

\subsection{Pipeline: \MDModel{}}
\begin{figure}[tb!]
\centering
\includegraphics[width=\textwidth]{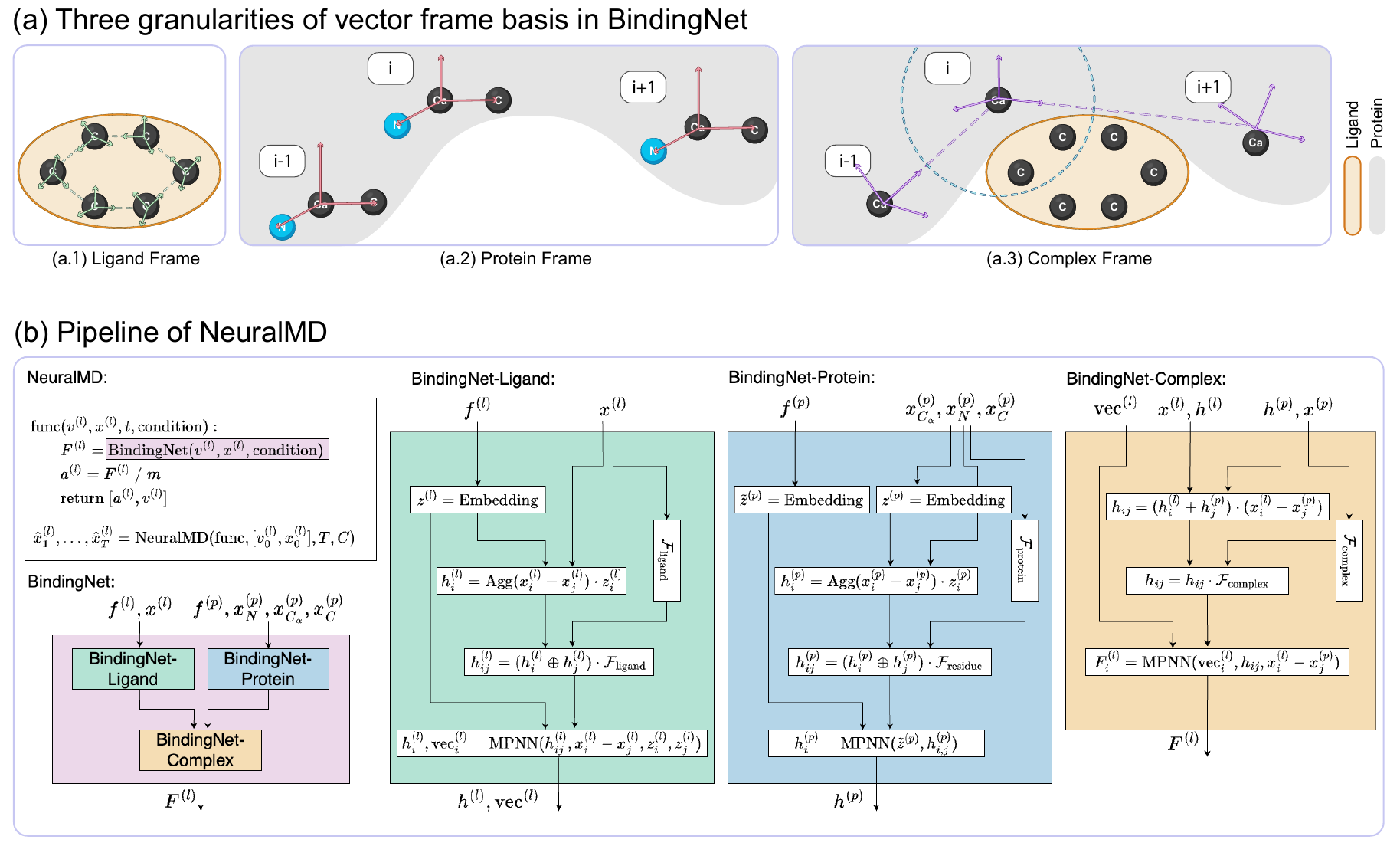}
\vspace{-4ex}
\caption{
(a) Three granularities of vector frame basis in \BindingNet{}.
(b) \BindingNet{} comprises three components and integrates into the pipeline of \MDModel{}.
}
\label{fig:vector_frame_and_NeuralMD_pipeline}
\vspace{-1ex}
\end{figure}

In this section, we briefly introduce \MDModel{}, our framework for learning molecular dynamics (MD) simulations in protein-ligand binding. It has two main phases: (1) \BindingNet{}: a multi-grained SE(3)-equivariant geometric model that represents the protein-ligand complex across three granularities. It utilizes vector frame bases at the atom level for ligands, the backbone level for proteins, and the residue level for protein-ligand complexes. (2) Dynamics Solver: either a second-order ordinary differential equation (ODE) solver or a second-order stochastic differential equation (SDE), used to model trajectories as Newtonian or Langevin dynamics, respectively. \Cref{fig:vector_frame_and_NeuralMD_pipeline} depicts the key components and pipeline of \MDModel{}. Please check the Methods Section for more details.

\textbf{(1) Multi-Grained Vector Frames}. We consider three data perspectives: atom level for ligands, backbone level for proteins, and residue level for protein-ligand complexes. These perspectives allow us to construct vector frames corresponding to each level, serving as reference bases for SE(3)-geometric modeling.

\textbf{(2) Multi-grained SE(3)-equivariant Binding Modeling: \BindingNet{}.} To model protein-ligand binding prediction, we introduce \BindingNet{}, which leverages the three vector frames established in the first step. The core idea is to achieve rotation-equivariant modeling by projecting vector variables onto the reference frames.

\textbf{(3) Molecular Dynamics Modeling of Binding Complexes: \MDModel{}.} Finally, we leverage the binding energy predictions from \BindingNet{} to simulate protein-ligand binding dynamics. This involves using a second-order ordinary differential equation (ODE) for Newtonian dynamics and a stochastic differential equation (SDE) for Langevin dynamics. These three steps collectively form the foundation of \MDModel{} for modeling binding dynamics.

\subsection{Experiment Setting}
\textbf{Datasets.}
One of the main bottlenecks of studying ML for molecular dynamics simulation in protein-ligand binding is insufficient data. Recently, the community has put more effort into gathering the datasets, and we consider MISATO in our work~\cite{siebenmorgen2024misato}. It is built on 16,972 experimental protein-ligand complexes extracted from the protein data bank (PDB)~\cite{berman2000protein}. Such data is obtained using X-ray crystallography, Nuclear Magnetic Resonance (NMR), or Cryo-Electron Microscopy (Cryo-EM), where systematic errors are unavoidable. This motivates the MISATO project, which utilizes semi-empirical quantum mechanics for structural curation and refinement, including regularization of the ligand geometry. For each protein-ligand complex, the trajectory comprises 100 snapshots in 8 nanoseconds under the fixed temperature and pressure. In Supplementary D, we list the basic statistics of MISATO, {\eg}, the number of atoms in small molecule ligands, and the number of residues in proteins.

\textbf{Baselines.}
Using ML for energy and force prediction, followed by trajectory prediction using numerical integration method, has been widely explored in the community, {\eg}, HDNNPs~\cite{behler2007generalized}, DeePMD~\cite{zhang2018deep}, TorchMD~\cite{doerr2020torchmd}, and Allegro-LAMMPS~\cite{musaelian2023scaling}. Here we extend this paradigm for binding dynamics and propose VerletMD, which utilizes \BindingNet{} for energy prediction on each snapshot and velocity Verlet algorithm to get the trajectory.
Additionally, we mainly focus on ML methods for trajectory prediction in this work, {\ie}, no energy or force is considered as labels. GNN-MD is to apply geometric graph neural networks (GNNs) to predict the trajectories in an auto-regressive manner~\cite{siebenmorgen2024misato,fu2022simulate}. More concretely, GNN-MD takes the coordinates and other molecular information as inputs at time $t$ and predicts the coordinates at time $t+1$. DenoisingLD (denoising diffusion for Langevin dynamics)~\cite{fu2022simulate,wu2023diffmd,arts2023two} is a baseline method that models the trajectory prediction as denoising diffusion task~\cite{song2020score}, and the inference for binding trajectory essentially becomes the Langevin dynamics. CG-MD learns a dynamic GNN and a score GNN~\cite{fu2022simulate}, which are essentially the hybrid of GNN-MD and DenoisingLD. Here, to make the comparison more explicit, we compare these two methods (GNN-MD and DenoisingLD) separately. We want to highlight that we keep the same backbone model, \BindingNet{}, for force or position prediction for all the baselines and \MDModel{}.

\textbf{Experiments Settings.}
We consider two experiment settings. The first type of experiment is the single-trajectory prediction, where both the training and test data are snapshots from the same trajectory, and they are divided temporally. The second type of experiment is the multi-trajectory prediction, where each data point is the sequence of all the snapshots from one trajectory, and the training and test data correspond to different sets of trajectories. All the experiments are conducted with three random seeds (0, 42, 123). The mean results are reported in the main article, and the standard deviations are reported in Supplementary F.\looseness=-1

\textbf{Evaluation Metrics.}
For quantitative evaluation, we use two reconstruction metrics and two validity metrics:
\begin{enumerate}[noitemsep,topsep=0pt]
    \item \textbf{Reconstruction.} For MD simulation, the evaluation is a critical factor for evaluating trajectory prediction. For both experiment settings, the trajectory reconstruction is the most straightforward evaluation metric. To evaluate this, we take both the mean absolute error (\textbf{MAE}) and mean squared error (\textbf{MSE}) between the predicted coordinates and ground-truth coordinates over all snapshots in the test set.
    \item \textbf{Validity.} Molecules possess certain inherent properties, and we would like to measure the validity of the sampled positions. The first validity metric is the \textbf{Matching} metric. It is defined as the mean squared error between the atom pairwise distance from ground-truth conformations and our sampled conformations. This Matching metric reveals whether the model correctly reflects the true distance relationships in physical space.
    The second validity metric is Stability. It is defined as $\mathbb{P}_{i, j} \big( \big\| \text{true distance} - \text{pred distance} \big\| \le \Delta \big)$, where we take $\Delta = 0.5$ \AA. The underlying intuition is that, over long-time simulations, the predicted trajectory could enter an unrealistic state, such as bond breaking. Stability quantifies how often the predicted distances between atoms stay within a reasonable range of the true distances, indicating that the trajectory does not exhibit pathological behavior.
\end{enumerate}

\subsection{Generalization Among Multiple Trajectories}

\begin{table}[tb!]
\setlength{\tabcolsep}{5pt}
\fontsize{9}{9}\selectfont
\centering
\caption{
\small
Results on three multi-trajectory binding dynamics predictions.
Four evaluation metrics are considered: MAE (\AA, $\downarrow$), MSE ($\downarrow$), Matching($\downarrow$), and Stability (\%, $\uparrow$).
}
\label{tab:main_result_multi_trajectory}
\vspace{-2ex}
\begin{adjustbox}{max width=\textwidth}
\begin{tabular}{l rrrr rrrr rrrr}
\toprule

\multirow{2}{*}{Dataset} & \multicolumn{4}{c}{MISATO-100}& \multicolumn{4}{c}{MISATO-1000} & \multicolumn{4}{c}{MISATO-All}\\
\cmidrule(lr){2-5} \cmidrule(lr){6-9} \cmidrule(lr){10-13}
 & \multicolumn{2}{c}{Reconstruction} & \multicolumn{2}{c}{Validity}
 & \multicolumn{2}{c}{Reconstruction} & \multicolumn{2}{c}{Validity}
 & \multicolumn{2}{c}{Reconstruction} & \multicolumn{2}{c}{Validity}
 \\
\cmidrule(lr){2-3} \cmidrule(lr){4-5}
\cmidrule(lr){6-7} \cmidrule(lr){8-9}
\cmidrule(lr){10-11} \cmidrule(lr){12-13}
 & MAE & MSE & Matching & Stability
 & MAE & MSE & Matching & Stability
 & MAE & MSE & Matching & Stability
 \\
\midrule
VerletMD & 85.286 & 54.996 & 46.753 & 10.051 & 104.537 & 68.942 & 48.899 & 10.574 & 97.213 & 64.405 & 50.857 & 11.888\\
GNN-MD & 5.964 & 3.938 & 0.671 & 70.546 & 7.524 & 4.915 & 0.670 & 68.310 & 7.637 & 5.048 & 0.675 & 69.244\\
DenoisingLD & 8.251 & 5.541 & 1.744 & 29.545 & 9.251 & 6.074 & 1.362 & 37.289 & 8.149 & 5.387 & 0.764 & 68.315\\
\midrule
\MDModel{} ODE (ours) & \textbf{5.867} & \textbf{3.870} & 0.539 & 79.553 & \textbf{7.459} & \textbf{4.867} & 0.612 & 70.362 & \textbf{7.513} & \textbf{4.961} & 0.491 & 81.991\\
\MDModel{} SDE (ours) & 5.868 & 3.871 & \textbf{0.533} & \textbf{80.229} & 7.476 & 4.876 & \textbf{0.457} & \textbf{83.960}  & 7.517 & 4.963 & \textbf{0.474} & \textbf{83.264}\\
\bottomrule
\end{tabular}
\end{adjustbox}
\vspace{-1ex}
\end{table}

The first task is to test the generalization ability of \MDModel{} among different trajectories. The MISATO dataset includes 13,765 protein-ligand complexes~\cite{siebenmorgen2024misato}, and we first create two small datasets by randomly sampling 100 and 1k complexes, respectively. Then, we take 80\%-10\%-10\% for training, validation, and testing on both datasets. We also consider the whole MISATO dataset, where the data split has already been provided. After removing the complexes with peptide ligands, we have 13,066, 1,357, and 1,357 complexes for training, validation, and testing, respectively.

The quantitative results are in~\Cref{tab:main_result_multi_trajectory}. The first observation is that we can observe that VerletMD has worse performance on all three datasets, and the performance gap with other methods is even larger compared to the single-trajectory prediction as will be introduced below. When comparing VerletMD, \MDModel{} reaches up to 15$\times$ reduction in reconstruction error and 70\% increase in validity. The second observation is that the other two baselines, GNN-MD and DenoisingLD, show similar performance, while \MDModel{} outperforms in all datasets and all four metrics. The last observation is that the two validity metrics (Matching and Stability) are more distinguishable than the two trajectory reconstruction metrics (MAE and MSE), suggesting that ligand oscillation may contain more informative details. We will explore this further in a qualitative study below.\looseness=-1

\begin{table}[!tb]
\setlength{\tabcolsep}{5pt}
\fontsize{9}{9}\selectfont
\centering
\caption{
\small
Results on ten single-trajectory binding dynamics predictions in the semi-flexible setting.
Results with optimal training loss are reported.
Four evaluation metrics are considered: MAE (\AA, $\downarrow$), MSE ($\downarrow$), Matching($\downarrow$), and Stability (\%, $\uparrow$).
}
\label{tab:main_result_single_trajectory}
\vspace{-2ex}
\begin{subtable}[c]{.49\textwidth}
\centering
\begin{adjustbox}{max width=\textwidth}
\begin{tabular}{l l rrr rr}
\toprule
PDB ID  & Metric & VerletMD & GNN-MD & DenoisingLD & \makecell{NeuralMD\\ODE (Ours)} & \makecell{NeuralMD\\SDE (Ours)}\\
\midrule
\multirow{4}{*}{5WIJ} & MAE & 14.629 & 2.280 & 2.501 & \textbf{2.252} & 2.260\\
 & MSE & 10.221 & 1.521 & 1.644 & \textbf{1.514} & \textbf{1.514}\\
 & Matching & 5.459 & 0.803 & 0.815 & \textbf{0.464} & 0.615\\
 & Stability & 24.360 & 54.475 & 52.418 & \textbf{82.046} & 67.464\\
\midrule
\multirow{4}{*}{4ZX0} & MAE & 21.278 & 2.370 & 3.138 & \textbf{1.878} & 2.158\\
 & MSE & 14.357 & 1.599 & 2.045 & \textbf{1.263} & 1.455\\
 & Matching & 7.971 & 0.555 & 1.072 & \textbf{0.428} & 0.696\\
 & Stability & 19.168 & 68.613 & 44.228 & \textbf{81.401} & 59.109\\
\midrule
\multirow{4}{*}{3EOV} & MAE & 27.960 & 3.512 & 4.055 & \textbf{3.858} & 3.395\\
 & MSE & 18.821 & 2.413 & 2.787 & \textbf{2.651} & 2.309\\
 & Matching & 13.588 & 1.216 & 1.209 & 1.062 & \textbf{0.962}\\
 & Stability & 13.067 & 40.984 & 41.469 & 47.328 & \textbf{50.108}\\
\midrule
\multirow{4}{*}{4K6W} & MAE & 15.428 & 3.695 & 3.942 & \textbf{3.656} & 3.765\\
 & MSE & 10.357 & 2.402 & 2.635 & \textbf{2.400} & 2.501\\
 & Matching & 7.505 & 1.038 & \textbf{0.839} & 0.928 & 1.076\\
 & Stability & 15.441 & 42.480 & \textbf{53.820} & 49.438 & 49.700\\
\midrule
\multirow{4}{*}{1KTI} & MAE & 18.157 & 6.641 & 7.051 & 6.675 & \textbf{6.646}\\
 & MSE & 12.723 & 4.173 & 4.369 & 4.176 & \textbf{4.141}\\
 & Matching & 7.467 & 0.386 & 0.268 & 0.337 & \textbf{0.167}\\
 & Stability & 19.352 & 81.831 & 91.986 & 86.430 & \textbf{98.508}\\
\bottomrule
\end{tabular}
\end{adjustbox}
\end{subtable}
\hfill
\begin{subtable}[c]{.49\textwidth}
\centering
\begin{adjustbox}{max width=\textwidth}
\begin{tabular}{l l rrr rr}
\toprule
PDB ID  & Metric & VerletMD & GNN-MD & DenoisingLD & \makecell{NeuralMD\\ODE (Ours)} & \makecell{NeuralMD\\SDE (Ours)}\\
\midrule
\multirow{4}{*}{1XP6} & MAE & 13.753 & 2.378 & 2.218 & \textbf{1.924} & 2.061\\
 & MSE & 9.587 & 1.561 & 1.472 & \textbf{1.280} & 1.356\\
 & Matching & 4.672 & 0.966 & 0.676 & \textbf{0.537} & 0.615\\
 & Stability & 28.129 & 49.239 & 64.951 & \textbf{75.533} & 69.423\\
\midrule
\multirow{4}{*}{4YUR} & MAE & 16.764 & 7.031 & 7.128 & \textbf{6.957} & 7.038\\
 & MSE & 11.069 & 4.641 & 4.807 & \textbf{4.597} & 4.679\\
 & Matching & 9.555 & 0.920 & 0.834 & \textbf{0.584} & 0.749\\
 & Stability & 16.542 & 47.555 & 49.676 & \textbf{69.775} & 60.344\\
\midrule
\multirow{4}{*}{4G3E} & MAE & 5.111 & 2.709 & 3.588 & \textbf{2.191} & 2.345\\
 & MSE & 3.503 & 1.785 & 2.321 & \textbf{1.453} & 1.536\\
 & Matching & 3.388 & 0.893 & 1.069 & \textbf{0.505} & 0.521\\
 & Stability & 31.852 & 61.802 & 40.823 & \textbf{71.436} & 68.729\\
\midrule
\multirow{4}{*}{6B7F} & MAE & 31.934 & 4.136 & 4.431 & 3.921 & \textbf{3.842}\\
 & MSE & 22.168 & 2.768 & 3.047 & 2.652 & \textbf{2.601}\\
 & Matching & 21.691 & 1.194 & 0.672 & \textbf{0.459} & 0.741\\
 & Stability & 11.050 & 39.067 & 61.583 & \textbf{75.692} & 57.917\\
\midrule
\multirow{4}{*}{3B9S} & MAE & 19.473 & 2.578 & \textbf{2.811} & 3.039 & 3.132\\
 & MSE & 11.696 & 1.699 & \textbf{1.868} & 1.999 & 2.078\\
 & Matching & 0.923 & 1.414 & 0.472 & 0.659 & \textbf{0.444}\\
 & Stability & 57.801 & 49.306 & 71.852 & 76.065 & \textbf{77.801}\\
\bottomrule
\end{tabular}
\end{adjustbox}
\end{subtable}
\end{table}

\subsection{Generalization of One Single Trajectory}
The second paradigm is the generalization of one single trajectory. The datasets are from the same protein-ligand trajectory, with the first 80 snapshots for training and the last 20 for testing.

The quantitative results are in~\Cref{tab:main_result_single_trajectory}.
The first observation is that the baseline VertletMD has a clear performance gap compared to the other methods. There are two possible reasons: (1) Using ML models to predict the energy (or force) at each snapshot, and then using a numerical integration algorithm can fail in the long-time simulations; (2) ML for energy prediction methods require more data to train than the ML for coordinate prediction methods, thus they can perform worse in the low-data regime.
The second observation is that, in general, both variants of \MDModel{} demonstrate competitive performance across all 10 tasks and four metrics, except for the reconstruction metrics on 3B9S and the validity metrics on 4K6W. The overall generalization performance reveals the potential of \MDModel{} compared to existing ML baselines. Specifically, Stability (\%) serves as a distinctive factor in method comparisons, with the two variants of \MDModel{} outperforming on 9 tasks by up to 70\%.\looseness=-1

\subsection{Qualitative Analysis on Oscillation}

\begin{figure}[tb!]
\centering
\includegraphics[width=\textwidth]{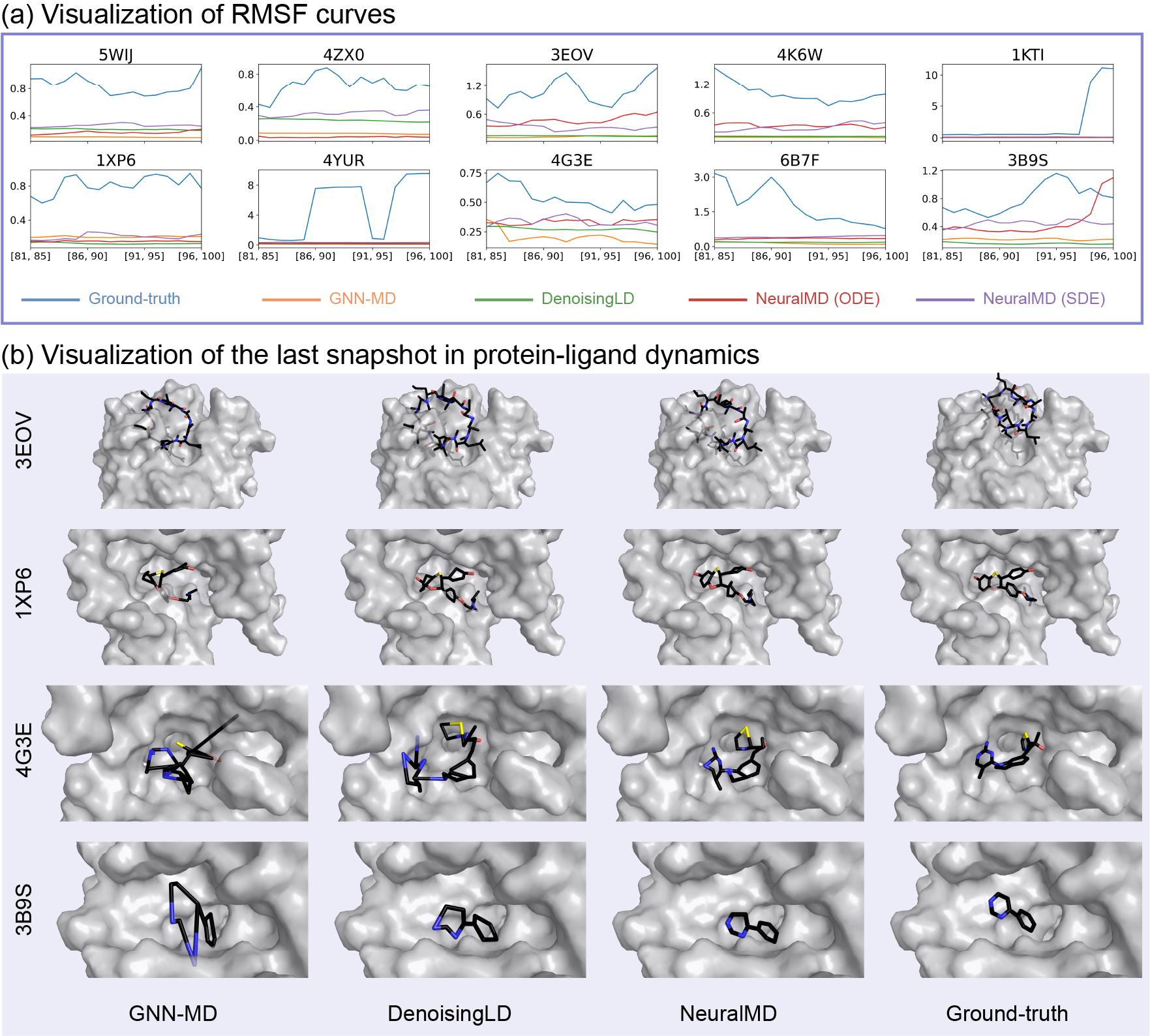}
\vspace{-4ex}
\caption{
(a) Oscillation visualization with RMSF.
(b) Visualization of the last snapshot on four protein-ligand dynamics. 
}
\label{fig:RMSF_and_trajectory}
\vspace{-1ex}
\end{figure}

In addition to the quantitative measures like reconstruction and validity, we would also like to discuss the qualitative performance of ML models for MD simulation.
We begin by analyzing the Root Mean Square Fluctuation (\textbf{RMSF}) along single trajectories. RMSF quantifies the mean square deviation of each atom's position from its average position over the trajectory, providing a measure of positional fluctuations. Formally, it is defined as:
\begin{equation}
\sqrt{\mathbb{E}_t \big[ \frac{1}{N}\sum_{i=1}^N \|r_{i,t} - \langle r_i \rangle \|^2} \big],
\end{equation}
where $r_{i,t}$ is the $i$-th atom position at time $t$ (ground truth or sampled by ML methods) and $\langle r_i \rangle$ is the average position for the $i$-th atom across time. To capture localized oscillations more effectively, we employ a sliding window approach, calculating RMSF over every 5 snapshots, such as snapshots [81, 85] and [96, 100].

RMSF is a metric that measures the flexibility of individual atoms or residues during a simulation, reflecting how much they deviate from their average positions over time. It provides valuable insights into the extent of fluctuations in the system. Specifically, RMSF describes the oscillations of individual trajectories. The objective here is to plot the RMSF curves for the ground truth and four ML methods, enabling a comparison to determine which ML method aligns most closely with the ground truth curve. Notice that VerletMD is not considered as it cannot converge on the single-trajectory experiments (according to ~\Cref{tab:main_result_single_trajectory}).
The RMSF curves are illustrated in \Cref{fig:RMSF_and_trajectory}. First, we observe that, overall, the two RMSF curves produced by \MDModel{} align better with the ground truth compared to those from ML baselines. Notably, \MDModel{} demonstrates a significant improvement on datasets PDB 4ZX0, 3EOV, 4K6W, 4G3E, and 3B9S, and shows slightly better performance on 5WIJ, 1XP6, and 6B7F. Second, we note that almost all ML methods perform poorly on 1KTI and 4YUR. For these two trajectories, the ground truth exhibits sudden positional changes, representing out-of-distribution movements that remain an open challenge for the current ML community.

In addition to RMSF, a more intuitive approach is to visualize the trajectories directly. We provide a 20-snapshot video comparing the trajectories generated by ML methods and the ground truth in the supplementary video. Additionally, we briefly present the sampled trajectories of four complexes using three ML methods alongside the ground truth in~\Cref{fig:RMSF_and_trajectory}b. From these visualizations, it is evident that GNN-MD occasionally collapses, while DenoisingLD maintains a comparatively structured trajectory. Notably, \MDModel{} demonstrates the highest stability across all cases.\looseness=-1

\subsection{Efficiency Analysis}

\begin{table}[tb!]
\setlength{\tabcolsep}{6.5pt}
\fontsize{9}{9}\selectfont
\centering
\caption{
\small
Efficiency comparison of FPS between VerletMD and \MDModel{} on single-trajectory prediction, with an accuracy of an integer.
}
\label{tab:main_table_computational_time}
\vspace{-2ex}
\begin{adjustbox}{max width=\textwidth}
\begin{tabular}{l | rrrrrrrrrr r}
\toprule
PDB ID & 5WIJ & 4ZX0 & 3EOV & 4K6W & 1KTI & 1XP6 & 4YUR & 4G3E & 6B7F & 3B9S & Average\\
\midrule
VerletMD & 516 & 529 & 485 & 427 & 439 & 483 & 510 & 520 & 540 & 542 & 499.1\\
GNN-MD & 830 & 806 & 763 & 715 & 756 & 768 & 823 & 810 & 823 & 844 & 793.8\\
DenoisingLD & 291 & 299 & 285 & 285 & 269 & 299 & 296 & 298 & 299 & 300 & 292.1\\
\MDModel{} ODE & 429 & 434 & 424 & 393 & 361 & 423 & 430 & 433 & 440 & 434 & 420.1\\
\MDModel{} SDE & 440 & 407 & 411 & 396 & 368 & 424 & 434 & 429 & 441 & 429 & 417.9\\
\bottomrule
\end{tabular}
\end{adjustbox}
\end{table}

Efficiency is another important metric, and we consider measuring it with frames per second (FPS) on a single Nvidia-V100 GPU card. One main benefit of using \MDModel{} for binding simulation is its efficiency. To show this, we list the computational time in~\Cref{tab:main_table_computational_time}. Since all ML methods share a similar backbone architecture (\BindingNet{} and its variants), their efficiency differences arise from their computational approaches. GNN-MD is the most efficient, as it does not require any integration steps. \MDModel{} incorporates an augmented integration step, with the optimal step size set to 0.5 of the snapshot interval, making it slightly slower than VerletMD. DenoisingLD is the slowest, as the optimal timestep between adjacent snapshots is 10.

We further approximate the wall time using the numerical method for MD simulation (PDB 5WIJ). Concretely, we can get an estimated speed of 1 nanosecond of dynamics every 0.28 hours. This is running the simulation with GROMACS~\cite{van2005gromacs} on 1 GPU with 16 CPU cores and a moderately sized water box at the all-atom level (around 64,000 atoms) with the stepsize of 2 femtoseconds. This shows that \MDModel{} is approximately 25K$\times$ faster than numerical methods under optimal conditions. However, as noted earlier, the model's efficiency is highly sensitive to hyperparameter choices, so we consider an efficiency improvement factor of at least 1K to be a conservative estimate.

\subsection{Case Study on 4G3E: Biological Meaning of \MDModel{}}
In this section, we provide a detailed discussion of PDB entry 4G3E, explore the potential advantages of using \MDModel{} for drug development, and examine its biological implications.

Nuclear factor kappa B (NF-$\kappa$B) is a type of transcriptional factor that regulates transcription, cell survival, immune responses, and the development of the immune system~\cite{liu2017nf,mulero2019genome,de2012crystal,park2016roles,hayden2011nf}. NF-$\kappa$B inducing kinase (NIK), also known as MAP3K14, is a crucial signaling molecule that participated in the non-canonical NF-$\kappa$B pathway, which regulates immune responses and lymphoid organ development~\cite{de2012crystal,hayden2011nf,thu2010nf}. NIK is a serine/threonine kinase that activates the pathway by phosphorylating IKK$\alpha$ (I$\kappa$B kinase $\alpha$), which in turn leads to the processing of NF-$\kappa$B2 (p100) into its active form, p52. This activation allows the p52-RelB complex to translocate to the nucleus and initiate gene transcription that regulates various immune functions, including B cell survival, osteoclastogenesis, and lymphoid tissue development~\cite{hayden2011nf,thu2010nf}.

NIK is a promising drug target due to its pivotal role in the non-canonical NF-$\kappa$B signaling pathway~\cite{de2012crystal,haselager2022therapeutic,crawford2023filling,pflug2020targeting}. Aberrant NIK activation and expression have been implicated in a range of pathological conditions, including autoimmune diseases, solid cancers, hematologic malignancies, cardiovascular disease, obesity, and type 2 diabetes~\cite{pflug2020targeting}. Consequently, small-molecule inhibitors targeting NIK have demonstrated therapeutic potential by modulating immune responses and reducing inflammation, making NIK an attractive target for the treatment of various diseases~\cite{crawford2023filling,pflug2020targeting}.

Many drugs targeting NIK have been developed, with several structures of NIK bound to inhibitors resolved, providing valuable insights for drug design~\cite{de2012crystal,cheng2021pharmacological,li2013inhibiting}. However, many existing drugs still require optimization to improve efficacy and specificity. Recent efforts used metabolite identification and structure-based drug design to improve pharmacokinetics, leading to inhibitors with reduced clearance and enhanced kinase selectivity, significantly lowering predicted human doses~\cite{crawford2023filling}. These advancements demonstrate how targeted optimization strategies can address key limitations. Our \MDModel{} approach can complement such strategies by offering dynamic insights into inhibitor binding, guiding the rational design of improved protein-targeted therapies.

The example of the complex structure of the inhibitor CMP1 bound to its target protein, NIK (Mus musculus; PDB: 4G3E), highlights the final binding state of the NIK inhibitor at its target region of NIK~\cite{de2012crystal}. However, the dynamic process leading to this interaction remains poorly understood. Despite the development of numerous approaches for simulating this inhibitor binding processes (\Cref{fig:RMSF_and_trajectory}), their effectiveness is limited. For example, these methods, GNN-MD and DenoisingLD, frequently produce unrealistic structural distortions of CMP1 during the simulation (\Cref{fig:RMSF_and_trajectory}).

In contrast, our \MDModel{} model not only accelerates the simulation of the binding trajectory between CMP1 and NIK but also avoids such exaggerated deformations. \MDModel{} simulations reveal that a specific structural region within the CMP1 molecule maintains stability or exhibits consistent, regular fluctuations during the binding process (\Cref{fig:RMSF_and_trajectory}b and supplementary video). These insights are invaluable for optimizing novel NIK inhibitors, particularly by targeting molecular regions that enhance binding affinity and selectivity.

Furthermore, similar results have been observed across numerous examples in our studies (supplementary video). This demonstrates the broad applicability of \MDModel{} for drug design, emphasizing its potential to streamline the development of novel therapeutics for various targets.

\section{Discussion}
To sum up, we devise \MDModel{}, an ML framework that incorporates a novel multi-grained group symmetric network architecture and second-order Newtonian mechanics, enabling accurate predictions of binding dynamics in a large timescale. Not only is such a timescale critical for understanding the dynamic nature of the ligand-protein complex, but our work marks the first approach in developing a framework to predict coordinates for MD simulation in protein-ligand binding. We quantitatively and qualitatively verify that \MDModel{} achieves superior performance on 13 binding dynamics tasks. 

One potential limitation of our work is the dataset. Currently, we are using the MISATO dataset, a protein-ligand dynamics dataset with a large timestep. However, \MDModel{} is agnostic to the timestep, and it can also be applied to binding dynamics datasets with timestep as a femtosecond. This may require the effort of the whole community for the dataset construction.

The second limitation of our work is GPU memory efficiency. Currently, we run multi- and single-trajectory MD simulations in a semi-flexible setting. While the flexible setting, where proteins can move, is more practical, it requires up to 100 times more GPU memory in the multi-trajectory case, as all protein conformations across 100 snapshots per trajectory must be modeled. For future work, we aim to develop a memory-efficient, physics-informed foundation model for proteins, capable of rapidly adapting to MD simulation tasks including MISATO.

Another limitation is the lack of verification through practical research. Although we have demonstrated the effectiveness of \MDModel{} through computational simulations, we have not yet validated its biological relevance using experimental techniques, such as X-ray crystallization and cryogenic electron microscopy. While our model has shown promising results in terms of simulating protein-ligand interactions, its applicability in drug development requires further confirmation through practical research. To address this gap, we plan to synthesize novel molecules based on the simulation results provided by \MDModel{}. Such studies will help to enhance the reliability of \MDModel{} as a tool for guiding the design of novel therapeutic agents.\looseness=-1

\section{Methods}
\subsection{Multi-Grained Vector Frames}
Proteins are macromolecules composed of up to thousands of residues (amino acids), where each residue is a small molecule. Thus, it is infeasible to model all the atoms in proteins due to the large volume of the system, and such an issue also holds for the protein-ligand complex. To address this issue, we propose \BindingNet{}, a multi-grained SE(3)-equivariant model, to capture the interactions between a ligand and a protein. The vector frame basis ensures SE(3)-equivariance, and the multi-granularity is achieved by considering frames at three levels.\looseness=-1

\textbf{Vector Frame Basis for SE(3)-Equivariance.}
Recall that the geometric representation of the whole molecular system needs to follow the physical properties of the equivariance w.r.t. rotation and translation. Such a group symmetric property is called SE(3)-equivariance. We also want to point out that the representation function should be reflection-equivariant for properties like energy, yet it is not for properties like chirality or ligand modeling in rigid protein structures. The vector frame basis inherently accommodates such reflection antisymmetry, and we leave a more detailed discussion in Supplementary C, along with the proof of group symmetry of the vector frame basis. In the following, we introduce three levels of vector frames for multi-grained modeling.

\textbf{Atom-Level Vector Frame for Ligands.}
For small molecule ligands, we first extract atom pairs $(i,j)$ within the distance cutoff $c$, and the vector frame basis is constructed using the Gram-Schmidt as:\looseness=-1
\begin{equation} \label{eq:atom_frame}
\small
\mathcal{F}_{\text{ligand}} = 
(\frac{\vx^{(l)}_{i} - \vx^{(l)}_{j}}{\norm{\vx^{(l)}_{i} - \vx^{(l)}_{j}}}, \frac{\vx^{(l)}_{i} \times \vx^{(l)}_{j}}{\norm{\vx^{(l)}_{i} \times \vx^{(l)}_{j}}},\frac{\vx^{(l)}_{i} - \vx^{(l)}_{j}}{\norm{\vx^{(l)}_{i} - \vx^{(l)}_{j}}} \times \frac{\vx^{(l)}_{i} \times \vx^{(l)}_{j}}{\norm{\vx^{(l)}_{i} \times \vx^{(l)}_{j}}}),
\end{equation}
where $\times$ is the cross product. Note that both $\vx^{(l)}_{i}$ and $\vx^{(l)}_{j}$ are for geometries at time $t$ - henceforth, we omit the subscript $t$ for brevity. Such an atom-level vector frame allows us to do SE(3)-equivariant message passing to get the atom-level representation.

\textbf{Backbone-Level Vector Frame for Proteins.}
Proteins can be treated as chains of residues, where each residue possesses a backbone structure. The backbone structure comprises an amino group, a carboxyl group, and an alpha carbon, delegated as $N-C_\alpha-C$. Such a structure serves as a natural way to build the vector frame. For each residue in the protein, the coordinates are $\vx_{N}$, $\vx_{C_\alpha}$, and $\vx_C$, then the backbone-level vector frame for this residue is:\looseness=-1
\begin{equation} \label{eq:protein_backbone_frame}
\mathcal{F}_{\text{protein}} = (
\frac{\vx_{N} - \vx_{C_\alpha}}{\norm{\vx_{N} - \vx_{C_\alpha}}},
\frac{\vx_{C_\alpha} - \vx_{C}}{\norm{\vx_{C_\alpha} - \vx_{C}}},
\frac{\vx_{N} - \vx_{C_\alpha}}{\norm{\vx_{N} - \vx_{C_\alpha}}} \times \frac{\vx_{C_\alpha} - \vx_{C}}{\norm{\vx_{C_\alpha} - \vx_{C}}}
).
\end{equation}
This is built for each residue, enabling the message passing for a residue-level representation.

\textbf{Residue-Level Vector Frame for Protein-Ligand Complexes.}
It is essential to model the protein-ligand interaction to better capture the binding dynamics. We achieve this by introducing the residue-level vector frame. More concretely, proteins are sequences of residues, marked as $\{(f^{(p)}_{0}, \vx^{(p)}_{0}), ..., (f^{(p)}_{i}, \vx^{(p)}_{i}), (f^{(p)}_{i+1}, \vx^{(p)}_{i+1}), ...\}$. Here, we use a cutoff threshold $c$ to determine the interactions between ligands and proteins, and the interactive regions on proteins are called pockets. We construct the following vector frame on residues in the pockets sequentially:
\begin{equation} \label{eq:complex_frame}
\mathcal{F}_{\text{complex}} = 
(\frac{\vx^{(p)}_{i} - \vx^{(p)}_{i+1}}{\norm{\vx^{(p)}_{i} - \vx^{(p)}_{i+1}}}, \frac{\vx^{(p)}_{i} \times \vx^{(p)}_{i+1}}{\norm{\vx^{(p)}_{i} \times \vx^{(p)}_{i+1}}},\frac{\vx^{(p)}_{i} - \vx^{(p)}_{i+1}}{\norm{\vx^{(p)}_{i} - \vx^{(p)}_{i+1}}} \times \frac{\vx^{(p)}_{i} \times \vx^{(p)}_{i+1}}{\norm{\vx^{(p)}_{i} \times \vx^{(p)}_{i+1}}} ).
\end{equation}
Through this vector frame, the message passing enables the exchange of information between atoms from ligands and residues from the pockets. The illustration of the above three levels of vector frames can be found in~\Cref{fig:vector_frame_and_NeuralMD_pipeline}. Once we build up such three vector frames, we then design \BindingNet{}, as will be introduced next. The key step involving vector frames is \textit{scalarization} operation~\cite{hsu2002stochastic}, which transforms the equivariant variables ({\eg}, coordinates) to invariant variables by projecting them to the three vector bases in the vector frame.

\subsection{Multi-Grained SE(3)-Equivariant Binding Force Modeling: \BindingNet{}}
In this section, we introduce \BindingNet{}, a multi-grained SE(3)-equivariant geometric model for protein-ligand binding. The input of \BindingNet{} is the geometry of the rigid protein and the ligand at time $t$, while the output is the force on each atom in the ligand.

\textbf{Atom-Level Ligand Modeling.}
We first generate the atom embedding using one-hot encoding and then aggregate each atom's embedding, $\vz^{(l)}$, by aggregating all its neighbor's embedding within the cutoff distance $c$. Then, we obtain the atom's equivariant representation by aggregating its neighborhood's messages as $(\vx^{(l)}_i - \vx^{(l)}_j) \cdot \vz^{(l)}_i$. A subsequent scalarization is carried out based on the atom-level vector frame as $\vh^{(l)}_{ij} = (\vh^{(l)}_i \oplus \vh^{(l)}_j) \cdot \mathcal{F}_{\text{ligand}}$, where $\oplus$ is the concatenation. Finally, it is passed through several equivariant message-passing layers (MPNN) defined as:
\begin{equation} \label{eq:MPNN_ligand}
\text{vec}^{(l)}_{i} = \text{vec}^{(l)}_i + \text{Agg}_{j} \big( \text{vec}^{(l)}_i \cdot \text{MLP}(\vh_{ij}) + (\vx^{(l)}_i - \vx^{(p)}_j) \cdot \text{MLP}(\vh_{ij}) \big),
\end{equation}
where $\text{MLP}(\cdot)$ and $\text{Agg}(\cdot)$ are the multi-layer perceptron and mean aggregation functions, respectively. vec $\in \mathbb{R}^3$ is a vector assigned to each atom and is initialized as 0. The outputs are atom representation and vector ($\vh^{(l)}$ and $\text{vec}^{(l)}$), and they are passed to the complex module introduced below.

\textbf{Backbone-Level Protein Modeling.}
For the coarse-grained modeling of proteins, we consider three backbone atoms in each residue. We first obtain the atom embedding on three atom types, and then we obtain each atom's representation $\vz^{(p)}$ by aggregating its neighbor's representation. Then, we obtain an equivariant atom representation by aggregating the edge information, $(\vx^{(p)}_i - \vx^{(p)}_j) \cdot \vz^{(p)}_i$, within cutoff distance $c$. Following which is the scalarization on the residue frame $\vh^{(p)}_{ij} = (\vh^{(p)}_i \oplus \vh^{(p)}_j) \cdot \mathcal{F}_{\text{protein}}$. Recall that we also have the residue type, and with a type embedding $\tilde \vz^{(p)}$, we can obtain the final residue-level representation using an MPNN layer as: 
\begin{equation} \label{eq:MPNN_protein}
\vh^{(p)} = \tilde \vz^{(p)} + (\vh^{(p)}_{N,C_\alpha} + \vh^{(p)}_{C_\alpha, C})/2.
\end{equation}

\textbf{Residue-Level Complex Modeling.}
Once we obtain the atom-level representation and vector ($\vh^{(l)}, \text{vec}^{(l)}$) from ligands, and backbone-level representation ($\vh^{(p)}$) from proteins, the next step is to learn the protein-ligand interaction. We first extract the residue-atom pair ($i, j$) with a cutoff $c$, based on which we obtain an equivariant interaction edge representation $\vh_{ij} = (\vh^{(l)}_i + \vh^{(p)}_j) \cdot (\vx^{(l)}_i - \vx^{(p)}_j)$. After scalarization, we can obtain invariant interaction edge representation $\vh_{ij} = \vh_{ij} \cdot \mathcal{F}_{\text{complex}}$. Finally, we adopt an equivariant MPNN layer to get the atom-level force as:
\begin{equation} \label{eq:MPNN_complex}
\text{vec}^{(pl)}_{ij} = \text{vec}^{(l)}_i \cdot \text{MLP}(\vh_{ij}) + (\vx^{(l)}_i - \vx^{(p)}_j) \cdot \text{MLP}(\vh_{ij}).
\end{equation}
The ultimate force predictions for each atom include two parts: the internal force from the molecule $\text{vec}^{(l)}_i$ and the external force from the protein-ligand interaction $\text{vec}^{(pl)}_{ij}$. To sum them up, we have:
\begin{equation} \label{eq:force_prediction}
F^{(l)}_i = \text{vec}^{(l)}_i + \text{Agg}_{j \in \mathcal{N}(i)} \text{vec}^{(pl)}_{ij}.
\end{equation}
These three modules consist \BindingNet{}. More complete descriptions can be found in Supplementary E.

\subsection{Binding Molecular Dynamics Modeling: \MDModel{}}
As previously clarified, molecular dynamics follows Newtonian mechanics, and we solve it as either an ODE problem or an SDE problem. The \BindingNet{} introduced above takes in the molecular system geometry at time $t$ and outputs the forces. Then in this section, we describe how we use the neural differential equation solver to predict the coordinates at future snapshots.

We want to highlight that one ML for the MD simulation research line is predicting the energy or force~\cite{doerr2020torchmd,musaelian2023scaling,zhang2018deep}, which will be fed into the numerical integration algorithms for trajectory simulation. For accuracy, such an ML-based MD simulation must be at the femtosecond level (1e-15 second). However, as shown in recent works~\cite{unke2021machine,stocker2022robust,fu2022forces}, minor errors in the ML force field can lead to catastrophic failure for long-time simulations. For instance, there can be pathological behavior, such as extreme force predictions or bond breaking, within the low end of the distribution. Our experiments have yielded similar observations, as will be shown below. In this paper, however, we overcome this issue by directly learning the extended-timescale MD trajectories (nanosecond level, 1e-9 second).

\textbf{Newtonian dynamics and Langevin dynamics for MD simulation.}
For MD simulation, if we assume the information of all particles in the molecular system ({\eg}, solvent molecules) is known and no thermal fluctuation is considered, then it can be modeled as Newtonian dynamics, which is essentially an ODE. For modeling, we consider using the \BindingNet{} introduced above for force prediction at time $\tau$, as:
\begin{equation} \label{eq:force_prediction_ODE}
F^{(l)}_{\tau}\text{-ODE} = \text{\BindingNet{}}(f^{(l)}, \vx^{(l)}_{\tau}, f^{(p)}, \vx^{(p)}_N, \vx^{(p)}_{C_\alpha}, \vx^{(p)}_C).
\end{equation}
On the other hand, Langevin dynamics introduces a stochastic component for large molecular systems with thermal fluctuations. Concretely, Langevin dynamics is an extension of the standard Newtonian dynamics with the addition of damping and random noise terms: $F - \gamma m \vv + \sqrt{2m \gamma k_BT} R(t)$, where $\gamma$ is the damping constant or collision frequency, $T$ is the temperature, $k_B$ is the Boltzmann's constant, and $R(t)$ is a delta-correlated stationary Gaussian process with zero-mean. For our implementation, we learn this equation and arguments in a data-driven way with reparameterization as:
\begin{equation} \label{eq:force_prediction_SDE}
F^{(l)}_{\tau}\text{-SDE} = \text{\BindingNet{}}(f^{(l)}, \vx^{(l)}_{\tau}, f^{(p)}, \vx^{(p)}_N, \vx^{(p)}_{C_\alpha}, \vx^{(p)}_C) + \text{\BindingNet{}-Ligand}(f^{(l)}, \vx^{(l)}_{\tau}) \cdot \epsilon,
\end{equation}
where $\epsilon \sim \mathcal{N}(0, 1)$ is the standard Gaussian noise. Ultimately, adopting either $F^{(l)}_{\tau}\text{-ODE}$ or $F^{(l)}_{\tau}\text{-SDE}$ as the force prediction $F^{(l)}_{\tau}$ on each atom, the coordinates at time $t$ can be obtained after integration as:
\begin{equation} \label{eq:second_order_integration}
\begin{aligned}
\va^{(l)}_{\tau} & = \frac{F^{(l)}_{\tau}}{m}, \qquad\qquad&
\hat \vv^{(l)}_{t} & = \vv^{(l)}_{0} + \int_0^t \va^{(l)}_{\tau} d \tau, \qquad\qquad&
\hat \vx^{(l)}_{t} & = \vx^{(l)}_{0} + \int_0^t \hat \vv^{(l)}_{\tau} d \tau.
\end{aligned}
\end{equation}
The training objective is the mean absolute error (MAE) or mean squared error (MSE) between the predicted coordinates and ground-truth coordinates along the whole trajectories:
\begin{equation} \label{eq:objective_function}
\mathcal{L} = \mathbb{E}_{t} \big[ ||\hat \vx^{(l)}_{t} - \vx^{(l)}_{t}|| \big].
\end{equation}
An illustration of \MDModel{} pipeline is in~\Cref{fig:vector_frame_and_NeuralMD_pipeline}.

\textbf{Second-order differential equation.}
We also want to highlight that \MDModel{} is solving a second-order differential equation, as in Newtonian and Langevin dynamics. The key module of the neural differential method is the differential function~\cite{chen2018neural}, which returns the first-order derivative. Then, the outputs of the differential function will be integrated using algorithms like the Euler algorithm. To learn the MD trajectory following second-order ODE and SDE, we propose the following formulation of the second-order equation within one integration call:
\begin{equation}\label{eq:second_order_integration_for_NeuralMD}
\begin{aligned}
\begin{bmatrix}
d \vx/dt\\
d \vv/dt
\end{bmatrix}
= 
\begin{bmatrix}
\vv\\
F / m
\end{bmatrix}.
\end{aligned}
\end{equation}
We mark this as ``func'' in~\Cref{fig:vector_frame_and_NeuralMD_pipeline}. This means we can augment ODE or SDE derivative space by concurrently calculating the accelerations and velocities, allowing simultaneous integration of velocities and positions. 

\textbf{Implementation details on velocity modeling.}
As in~\Cref{eq:second_order_integration,eq:second_order_integration_for_NeuralMD}, \MDModel{} requires both initial coordinates and velocities to predict the trajectories of coordinates and velocities. However, the MISATO dataset we are using does not include velocity information.
To handle this issue, we introduce two solutions:
(1) We first estimate the initial velocity as the positional momentum, $\vv^{(l)}_t = (\vx^{(l)}_{t} - \vx^{(l)}_{t-1})$. Optionally, we can apply a velocity initial mapping function to get a surrogate initial velocity: $\vv^{(l)}_t = \text{\BindingNet{}-Ligand}(f^{(l)}, \vx^{(l)}_{t} - \vx^{(l)}_{t-1})$. The decision to include a velocity mapping is treated as a binary hyperparameter.
(2) During the integration step in~\Cref{eq:second_order_integration_for_NeuralMD}, we can also take an extra module to refine the velocity. More specifically, we have $\vv^{(l)}_t = \vv^{(l)}_t + \alpha \cdot \text{\BindingNet{}-Ligand}(f^{(l)}, \vx^{(l)}_t)$, where $\alpha$ is the velocity refinement coefficient.
This is not displayed from~\Cref{fig:vector_frame_and_NeuralMD_pipeline} for brevity, yet all critical details are provided to ensure reproducibility.

\subsection{Summary}
To sum up, we have introduced the three levels of vector frame (\Cref{eq:atom_frame,eq:protein_backbone_frame,eq:complex_frame}), based on which we introduce \BindingNet{}, a multi-grained SE(3)-equivariant geometric model for protein-ligand binding force prediction (\Cref{eq:MPNN_ligand,eq:MPNN_protein,eq:MPNN_complex,eq:force_prediction}). Finally, we take \BindingNet{} to get the force prediction (\Cref{eq:force_prediction_ODE,eq:force_prediction_SDE}), which will be then injected into the second-order integration to get the position prediction along the trajectory (\Cref{eq:second_order_integration,eq:second_order_integration_for_NeuralMD} and \Cref{fig:vector_frame_and_NeuralMD_pipeline}). The objective function is the MAE or MSE on position prediction (\Cref{eq:objective_function}). The algorithm is presented in~\Cref{algo:NeuralMD}.

\noindent
\begin{minipage}{\textwidth}
\begin{algorithm}[H]
\fontsize{8.2}{1}\selectfont
\caption{\small Training of \MDModel{}} \label{algo:NeuralMD}
\begin{algorithmic}[1]
    \STATE {\bfseries Input:} Initial position $\vx^{(l)}_{0}$ and initial velocity $\vv^{(l)}_{0}$ for ligands, atomic features $f^{(l)}$ for ligands, residue types and coordinates $f^{(p)}, \vx^{(p)}_{N}, \vx^{(p)}_{C_\alpha}, \vx^{(p)}_{C}$ for proteins, and time $T$.
    \FOR{each batch of binding trajectories}
    \FOR{discretized time $t \in \{1, 2, ..., T-1\}$}
        \STATE Centralize the coordinates of the ligand-protein complex for $\vx^{(l)}_t$ and $\vx^{(p)}$ by removing the trajectory mass center.
        \STATE Set T\_list = $[t, t+1, ...]$ and condition $C=[f^{(l)}, f^{(p)}, \vx^{(p)}_{N}, \vx^{(p)}_{C_\alpha} ,\vx^{(p)}_{C}]$.
        \STATE Get predicted position $[\hat \vx^{(l)}_{t+1}, \hat \vx^{(l)}_{t+2}, ...] = \text{\MDModel{}}(\text{func}, [\vx^{(l)}_t, \vv^{(l)}_t], \text{T\_list}, C)$.\hfill // \Cref{eq:second_order_integration_for_NeuralMD}
        \STATE Calculate the position prediction loss $\mathcal{L} = \mathbb{E}_{t} \big[ |\hat \vx^{(l)}_t - \vx^{(l)}_t | \big]$.\hfill // \Cref{eq:objective_function}
        \STATE Update \BindingNet{} parameters to minimize $\mathcal{L}$.
    \ENDFOR
    \ENDFOR
\end{algorithmic}
\end{algorithm}
\end{minipage}

\section*{Code Availability}
The code is available on this \href{https://github.com/chao1224/NeuralMD}{GitHub repo}.

\section*{Data Availability}
The data utilized in this work is available at MISATO~\cite{siebenmorgen2024misato}. Detailed instructions for downloading and preprocessing the dataset are provided at this  \href{https://github.com/chao1224/NeuralMD}{GitHub repo}.

\section*{Author Contributions Statement}
All authors contributed to the project discussion and paper writing.
S.L., W.D., H.X., Y.L., Z.L., V.B., C.B., A.A., H.G., and J.C. conceived and designed the experiments.
S.L., W.D., and Y.L. performed the experiments.
S.L., W.D., H.X., Y.L., Z.L., V.B., and D.Y. analyzed the data.
S.L., H.X., Y.L., Z.L., V.B., and D.Y. contributed analysis tools.

\printbibliography[title={References}]

\newpage
\appendix

\section{Preliminaries and Related Works}

\subsection{Preliminaries on Molecular Dynamics}
Molecular dynamics (MD) simulations predict how every atom in a molecular system moves over time, which is determined by the interatomic interactions following Newton's second law. Such a molecular system includes small molecules~\cite{zhang2018deep,rapaport2004art}, proteins~\cite{arts2023two,karplus2005molecular,frauenfelder2009unified}, polymers~\cite{fu2022simulate}, crystals~\cite{liu2024CrystalFlow}, and protein-ligand complexes~\cite{siebenmorgen2024misato,korlepara2022plas}. Typically, an MD simulation is composed of two main steps, {\ie}, (1) the energy and force calculation and (2) integration of the equations of motion governed by Newton’s second law of motion, using the initial conditions and forces calculated in step (1). As the initial condition, the initial positions and velocities are given for all the particles ({\eg}, atoms) in the molecular system; the MD simulation repeats the two steps to get a trajectory. Such MD simulations can be used to calculate the equilibrium and transport properties of molecules, materials, and biomolecular systems~\cite{frenkel2002understanding}.

In such an MD simulation, one key factor is estimating the forces on each atom. The function that gives the energy of a molecular system as a function of its structure (and forces via the gradient of the energy with respect to those atomic coordinates) is referred to as a potential energy surface (PES). In general, MD simulations integrate the equations of motion using a PES from one of two sources: (1) Classical MD using the force fields, which are parameterized equations that approximate the true PES, and are less costly to evaluate, allowing for the treatment of larger systems and longer timesteps. (2) \textit{ab-inito} MD (which calculates the energy of a molecular system via electronic structure methods, {\eg}, DFT) provide more accurate PES, but are limited in the system size and timesteps that are practically accessible due to the cost of evaluating the PES at a given point.

\subsection{Related Works}
\textbf{SE(3)-Equivariant Representation for Small Molecules and Proteins.}
The molecular systems are indeed a set of atoms located in the 3D Euclidean space. From a machine learning point of view, the representation function or encoding function of such molecular systems needs to be group-symmetric, {\ie}, the representation needs to be equivariant when we rotate or translate the whole system. Such symmetry is called the SE(3)-equivariance. Recently works~\cite{liu2023symmetry,zhang2023artificial} on molecules has provided a unified way of equivariant geometric modeling. They categorize the mainstream representation methods into three big venues: SE(3)-invariant models, SE(3)-equivariant models with spherical frame basis, and SE(3)-equivariant models with vector frame basis. (1) Invariant models that utilize invariant features (distances and angles) to predict the energies~\cite{schutt2018schnet,klicpera2020fast}, but the derived forces are challenging for ML optimization after integration. (2) Equivariant models with spherical frames that include a computationally expensive tensor-product operation~\cite{thomas2018tensor,musaelian2022learning}, which is unsuitable for large molecular systems. (3) Equivariant models with vector frames that have been explored for single stable molecules, including molecule representation and pretraining~\cite{liu2023group,du2023molecule,satorras2021n,schutt2021equivariant}, molecule conformation generation~\cite{du2022se}, protein representation~\cite{fan2022continuous}, and protein folding and design~\cite{ingraham2019generative,jumper2021highly}. However, no one has tried it for binding complexes.

\textbf{ML for Potential Energy and Force Learning for MD Simulation.}
One straightforward way of molecular dynamics (MD) simulation is through potential energy modeling. \textit{Numerical methods} for MD simulation can be classified into classical MD and \textit{ab-initio} MD, depending on using the classical mechanism or quantum mechanism to calculate the forces. Alternatively, a machine learning (ML) research line is to adopt geometric representation methods to learn the energies or the forces, {\eg}, by the geometric methods listed above. The first work is DeePMD~\cite{zhang2018deep}, which targets learning the potential energy function at each conformation. For inference, the learned energy can be applied to update the atom placement using i-PI software~\cite{ceriotti2014more}, composing the MD trajectories. TorchMD~\cite{doerr2020torchmd} utilizes TorchMD-Net~\cite{tholke2022equivariant} for energy prediction, which will be fed into the velocity Verlet algorithm for MD simulation. Similarly, \citeauthor{musaelian2023scaling} adopts Allegro~\cite{musaelian2023learning} model to learn the force at each conformation. The learned model will be used for MD trajectory simulation using LAMMPS~\cite{thompson2022lammps}. In theory, all the geometric models on small molecules~\cite{liu2023group,satorras2021n,schutt2021equivariant} and proteins~\cite{fan2022continuous} can be applied to the MD simulation task. However, there are two main challenges: (1) They require the interval between snapshots to be at the femtoseconds (1e-15 seconds) level for integration to effectively capture the motion of the molecules. (2) They take the position-energy pairs independently, and thus, they ignore their temporal correlations during learning.

\textbf{ML for Trajectory Learning for MD Simulation.}
More recent works have explored MD simulation by directly learning the coordinates along the trajectories. There are two key differences between energy and trajectory prediction for MD: (1) Energy prediction takes each conformation and energy as IID, while trajectory learning optimizes the conformations along the whole trajectory, enforcing the temporal relation. (2) The trajectory learning is agnostic to the magnitude of the timesteps, and energy prediction can be sensitive to longer-timestep MD simulations. More concretely, along such trajectory learning research line, CGDMS~\cite{greener2021differentiable} builds an SE(3)-invariant model, followed by the velocity Verlet algorithm for MD simulation. DiffMD~\cite{wu2023diffmd} is a Markovian method and treats the dynamics between two consecutive snapshots as a coordinate denoising process. It then applies the SDE solver~\cite{song2020score} to solve the molecular dynamics. A parallel work, DFF~\cite{arts2023two}, applies a similar idea for MD simulation. CG-MD~\cite{fu2022simulate} encodes a hierarchical graph neural network model for an auto-regressive position generation and then adopts the denoising method for fine-tuning. However, these works disregard the prior knowledge of the Newtonian mechanics governing the motion of atoms.

\begin{table}[t]
\centering
\caption{
\small Comparison of different numerical and machine learning (ML) methods for molecular dynamics (MD).
AR for autoregressive and denoising for denoising diffusion method.
}
\vspace{-2ex}
\begin{adjustbox}{max width=\textwidth}
\begin{tabular}{l l l l l l l}
\toprule
Category & Method & Energy / Force Calculation & Dynamics & Objective Function & Publications\\
\midrule
\multirow{6}{*}{Numerical Methods}
& Classical MD & \makecell[l]{Classical Mechanics:\\Force Field} & Newtonian Dynamics & -- & --\\
\cmidrule(lr){2-6}
& \textit{Ab-initio} MD & \makecell[l]{Quantum Mechanics:\\DFT for Schrodinger Equation} & Newtonian Dynamics & -- & --\\
\cmidrule(lr){2-6}
& Langevin MD & \makecell[l]{Classical Mechanics:\\Force Field} & Langevin Dynamics & -- & --\\
\midrule
\multirow{11}{*}{ML Methods}
& DeePMD~\cite{zhang2018deep} & Atom-level Modeling & Newtonian Dynamics (i-PI) & Energy Prediction & PRL'18\\
& TorchMD~\cite{doerr2020torchmd} & Atom-level Modeling & Newtonian Dynamics (velocity Verlet) & Energy Prediction & ACS'20\\
& Allegro-LAMMPS~\cite{musaelian2023scaling} & Atom-level Modeling & Newtonian Dynamics (LAMMPS) & Force Prediction & ArXiv'23
\\
& \textbf{VerletMD (Ours, baseline)} & Atom-level Modeling & Newtonian Dynamics (velocity Verlet) & Energy Prediction & --\\
\cmidrule(lr){2-6}
& CGDMS~\cite{greener2021differentiable} & Atom-level Modeling & Newtonian Dynamics (velocity Verlet) & Position Prediction & PLOS'21\\
& DiffMD~\cite{wu2023diffmd} & Atom-level Modeling & AR + Denoising & Position Prediction & AAAI'23\\
& DFF~\cite{arts2023two} & Atom-level Modeling & AR + Denoising & Position Prediction & ACS'23\\
& CG-MD~\cite{fu2022simulate} & Atom-level Modeling & AR + Denoising & Position Prediction & TMLR'23\\
& \textbf{LigandMD (Ours, baseline)} & Atom-level Modeling & AR + Denoising & Position Prediction & --\\
& \textbf{\MDModel{} ODE (Ours)} & Atom-level Modeling & Newtonian Dynamics & Position Prediction & --\\
& \textbf{\MDModel{} SDE (Ours)} & Atom-level Modeling & Langevin Dynamics & Position Prediction & --\\
\bottomrule
\end{tabular}
\end{adjustbox}
\end{table}

\textbf{MD Simulation in Protein-Ligand Binding.}
The MD simulation papers discussed so far are mainly for small molecules or proteins, not the binding dynamics. On the other hand, many works have studied the protein-ligand binding problem in the equilibrium state~\cite{stepniewska2018development,jimenez2018k,jones2021improved,yang2023geometric}, but not the dynamics. In this work, we consider a more challenging task, which is the protein-ligand binding dynamics. The viability of this work is also attributed to the efforts of the scientific community, where more binding dynamics datasets have been revealed, including PLAS-5k~\cite{korlepara2022plas}, MISATO~\cite{siebenmorgen2024misato}, and PLAS-20k~\cite{priyakumar2023plas}.

\clearpage
\newpage
\section{Group Symmetry and Equivariance}

In this article, a 3D molecular graph is represented by a collection of 3D point clouds. The corresponding symmetry group is SE(3), which consists of translations and rotations. Recall that we define the notion of equivariance functions in $\mathbf{R}^3$ in the main text through group actions. Formally, the group SE(3) is said to act on $\mathbf{R}^3$ if there is a mapping $\phi: \text{SE(3)} \times \mathbf{R}^3 \rightarrow \mathbf{R}^3$ satisfying the following two conditions:
\begin{enumerate}[noitemsep,topsep=0pt]
    \item if $e \in \text{SE(3)}$ is the identity element, then
    $$\phi(e,\vr) = \vr\ \ \ \ \text{for}\ \ \forall \vr \in \mathbf{R}^3.$$
    \item if $g_1,g_2 \in \text{SE(3)}$, then
    $$\phi(g_1,\phi(g_2,\vr)) = \phi(g_1g_2, \vr)\ \ \ \ \text{for}\ \ \forall \vr \in \mathbf{R}^3.$$
\end{enumerate}
Then, there is a natural SE(3) action on vectors $\vr$ in $\mathbf{R}^3$ by translating $\vr$ and rotating $\vr$ for multiple times. For $g \in \text{SE(3)}$ and $\vr \in \mathbf{R}^3$, we denote this action by $g\vr$. Once the notion of group action is defined, we say a function $f:\mathbf{R}^3 \rightarrow \mathbf{R}^3$ that transforms $\vr \in \mathbf{R}^3$ is equivariant if:
$$f(g\vr) = g f(\vr),\ \ \  \text{for}\ \ \forall\ \  \vr \in \mathbf{R}^3.$$
On the other hand, $f:\mathbf{R}^3 \rightarrow \mathbf{R}^1$ is invariant, if $f$ is independent of the group actions:
$$f(g\vr) = f(\vr),\ \ \  \text{for}\ \ \forall\ \  \vr \in \mathbf{R}^3.$$
For some scenarios, our problem is chiral sensitive. That is, after mirror reflecting a 3D molecule, the properties of the molecule may change dramatically. In these cases, it's crucial to include reflection transformations into consideration. More precisely, we say an SE(3) equivariant function $f$ is \textbf{reflection anti-symmetric}, if:
\begin{equation}
    \label{def: reflection antisymmetry}
f(\rho \vr) \neq f(\vr),
\end{equation}
for reflection $\rho \in \text{E(3)}$. 

\clearpage
\newpage
\section{Equivariant Modeling With Vector Frames}

\textbf{Frame} is a popular terminology in science areas. In physics, the frame is equivalent to a coordinate system. For example, we may assign a frame to all observers, although different observers may collect different data under different frames, the underlying physics law should be the same. In other words, denote the physics law by $f$, then $f$ should be an equivariant function.

There are certain ways to choose the frame basis, and below we introduce two main types: the orthogonal basis and the protein backbone basis. The orthogonal basis can be built for flexible 3D point clouds such as atoms, while the protein backbone basis is specifically proposed to capture the protein backbone.

\subsection{Basis} \label{sec:orthogonal_basis}
Since there are three orthogonal directions in $\mathbf{R}^3$, one natural frame in $\mathbf{R}^3$ can be a frame consisting of three orthogonal vectors:
$$F = (\ve_1,\ve_2,\ve_3).$$
Once equipped with a frame (coordinate system), we can project all geometric quantities to this frame. For example, an abstract vector $\vx \in \mathbf{R}^3$ can be written as $\vx = (r_1,r_2,r_3)$ under the frame
$F$, if:
$\vx = r_1 \ve_1 + r_2 \ve_2 + r_3 \ve_3.$
A vector frame further requires the three orthonormal vectors in $(\ve_1,\ve_2,\ve_3)$ to be equivariant. Intuitively, a vector frame will transform according to the global rotation or translation of the whole system. Once equipped with a vector frame, we can project vectors into this frame in an equivariant way:
\begin{equation} \label{projection}
\vx = \Tilde{r}_1 \ve_1 + \Tilde{r}_2 \ve_2 + \Tilde{r}_3 \ve_3.    
\end{equation}
We call the process of $\vx \rightarrow \Tilde{r}:= (\Tilde{r}_1,\Tilde{r}_2,\Tilde{r}_3)$ the \textbf{scalarization} or \textbf{projection} operation. Since $\Tilde{r}_i = \ve_i \cdot \vx$ is expressed as an inner product between vector vectors, we know that $\Tilde{r}$ consists of scalars. 

In this article, we assign a vector frame to each node/edge, therefore we call them the local frames. We want to highlight that, in this section, we prove the equivariance property of the vector frame basis using the Gram-Schmidt project. However, the similar equivariance property can be easily guaranteed for the vector frame bases in the main article after we remove the mass center of the molecular system.

In the main body, we constructed three vector frames based on three granularities. Here we provide the proof on the protein backbone frame. Say the three backbone atoms in on proteins are $\vx_i, \vx_j, \vx_k$ respectively. Then the vector frame is defined by:
\begin{equation} \label{def: edge frame}
\text{Vector-Frame}( \vx_i,\vx_j):= \textbf{Gram-Schmidt}\{ \vx_i -\vx_j, \vx_i - \vx_k, (\vx_i -\vx_j) \times (\vx_i - \vx_k)\}. 
\end{equation}
The Gram-Schmidt orthogonalization makes sure that the $\text{Vector-Frame}( \vx_i,\vx_j)$ is orthonormal.

\paragraph{Reflection Antisymmetric}
Since we implement the cross product $\times$ for building the local frames,  the third vector in the frame is a pseudo-vector. Then, the \textbf{projection} operation is not invariant under reflections (the inner product between a vector and a pseudo-vector change signs under reflection). Therefore, our model can discriminate two 3D geometries with different chiralities.

Our local frames also enable us to output vectors equivariantly by multiplying scalars $(v_1,v_2,v_3)$ with the frame: $\vv = v_1 \cdot \ve_1 + v_2 \cdot \ve_2 + v_3 \cdot \ve_3.$

\vspace{+2ex}
\noindent
\fbox{\parbox{\textwidth}{
\paragraph{Equivariance w.r.t. cross-product}
The goal is to prove that the cross-product is equivariant to the SE(3)-group, {\ie}:
\begin{equation}
\begin{aligned}
g x \times g y 
& = g (x \times y), \quad\quad g \in \text{SE(3)-Group}\\
\end{aligned}
\end{equation}
}}

\begin{proof}
\textbf{Geometric proof.}
From intuition, with rotation matrix $g$, we are transforming the whole basis, thus the direction of $g x \times g y$ changes equivalently with $g$. And for the value/length of $g x \times g y$, because $|g x \times g y| = \|g x \| \cdot \| g y \| \cdot \sin{\theta} = \|x \| \cdot \|y \| \cdot \sin{\theta} = |x \times y|$. So the length stays the same, and the direction changes equivalently. Intuitively, this interpretation is quite straightforward. 

\textbf{Analytical proof.} A more rigorous proof can be found below:

First, we have that for the rotation matrix $g$:
\begin{equation}
\begin{aligned}
g x \times g y
=
\begin{bmatrix} \vg_1^T \vx\\ \vg_2^T  \vx \\ \vg_3^T \vx \end{bmatrix}
\times 
\begin{bmatrix} \vg_1^T \vy \\ \vg_2^T \vy \\ \vg_3^T \vy \end{bmatrix} 
= 
\begin{bmatrix}
\vg_2^T \vx \cdot \vg_3^T \vy - \vg_3^T x \cdot \vg_2^T \vy\\
-\vg_1^T \vx \cdot \vg_3^T \vy + \vg_3^T \vx \cdot \vg_1^T \vy\\
\vg_1^T \vx \cdot \vg_2^T \vy - \vg_2^T \vx \cdot \vg_1^T \vy
\end{bmatrix},
\end{aligned}
\end{equation}
where $\vg_i, \vx, \vy \in \mathbb{R}^{3 \times 1}$.

Because $A^T C \cdot B^T D - A^TD \cdot B^T C = (A \times B)^T (C \times D)$, so we can have:
\begin{equation}
\begin{aligned}
g x \times g y
=
\begin{bmatrix}
\vg_2^T \vx \cdot \vg_3^T \vy - \vg_3^T x \cdot \vg_2^T \vy\\
-\vg_1^T \vx \cdot \vg_3^T \vy + \vg_3^T \vx \cdot \vg_1^T \vy\\
\vg_1^T \vx \cdot \vg_2^T \vy - \vg_2^T \vx \cdot \vg_1^T \vy
\end{bmatrix}
= 
\begin{bmatrix}
(\vg_2 \times \vg_3)^T (\vx \times \vy)\\
(\vg_3 \times \vg_1)^T (\vx \times \vy)\\
(\vg_1 \times \vg_2)^T (\vx \times \vy).
\end{bmatrix}
\end{aligned}
\end{equation}

Then because:
\begin{equation}
\begin{aligned}
& \text{det}(g) = (\vg_2 \times \vg_3)^T \vg_1 = \vg_1^T \vg_1 = 1\\
\Longrightarrow & (\vg_2 \times \vg_3)^T \vg_1 \vg_1^{-1} = \vg_1^T \vg_1 \vg_1^{-1} \\
\Longrightarrow & (\vg_2 \times \vg_3)^T  = \vg_1^T.\\
\end{aligned}
\end{equation}

Thus, we can have
\begin{equation}
\begin{aligned}
g x \times g y
= 
\begin{bmatrix}
(\vg_2 \times \vg_3)^T (\vx \times \vy)\\
(\vg_3 \times \vg_1)^T (\vx \times \vy)\\
(\vg_1 \times \vg_2)^T (\vx \times \vy)
\end{bmatrix}
= 
\begin{bmatrix}
\vg_1^T (\vx \times \vy)\\
\vg_2^T (\vx \times \vy)\\
\vg_3^T (\vx \times \vy)
\end{bmatrix}
= g (\vx \times \vy).
\end{aligned}
\end{equation}
\end{proof}

\vspace{+2ex}
\noindent
\fbox{\parbox{\textwidth}{
\paragraph{Rotation symmetric}
The goal is to prove

\begin{equation}
\begin{aligned}
\text{Vector-Frame}(g \vx_i, g \vx_j)
& = g  \textbf{Gram-Schmidt}\{ \vx_i -\vx_j, \vx_i - \vx_k, (\vx_i -\vx_j) \times (\vx_i - \vx_k)\}.
\end{aligned}
\end{equation}
}}

\begin{proof}
We can have:
\begin{equation}
\begin{aligned}
\text{Vector-Frame}(g \vx_i, g \vx_j)
& = \textbf{Gram-Schmidt}\{ g \vx_i - g \vx_j, g \vx_i - g \vx_k, (g \vx_i - g \vx_j) \times (g \vx_i - g \vx_k)\}\\
& = \textbf{Gram-Schmidt}\{ g (\vx_i - \vx_j), g (\vx_i - \vx_k), g ((\vx_i - \vx_j) \times (\vx_i -\vx_k))\}.
\end{aligned}
\end{equation}

Recall that Gram-Schmidt projection ($ \textbf{Gram-Schmidt}\{\vv_1, \vv_2, \vv_3\}$) is:
\begin{equation}
\begin{aligned}
& \vu_1 = \vv_1,
& \quad\quad\quad \ve_1 = \frac{\vv_1}{\| \vv_1 \|},\\
& \vu_2 = \vv_2 - \frac{\vu_1^T \vv_2}{\|\vu_1\|} \vu_1,
& \quad\quad\quad \ve_2 = \frac{\vv_2}{\| \vv_2 \|},\\
& \vu_3 = \vv_3 - \frac{\vu_1^T \vv_3}{\| \vu_1 \|} \vu_1 - \frac{\vu_2^T \vv_3}{\|\vu_2 \|} \vu_2,
& \quad\quad\quad \ve_3 = \frac{\vv_3}{\| \vv_3 \|}.
\end{aligned}
\end{equation}

Thus, the  Gram-Schmidt projection on the rotated vector ($ \textbf{Gram-Schmidt}\{\vg \vv_1, \vg \vv_2, \vg \vv_3\}$) is:
\begin{equation}
\begin{aligned}
& \vu_1' = \vg \vv_1,\\
%
& \vu_2' = \vg \vv_2 - \vg \frac{\vu_1^T \vv_2}{\|\vu_1\|} \vu_1,\\
%
& \vu_3' = \vg \vv_3 - \vg \frac{\vu_1^T \vv_3}{\| \vu_1 \|} \vu_1 - \vg \frac{\vu_2^T \vv_3}{\|\vu_2 \|} \vu_2,
\end{aligned}
\end{equation}

Thus, $\textbf{Gram-Schmidt}\{\vg \vv_1, \vg \vv_2, \vg \vv_3\} = g \textbf{Gram-Schmidt}\{\vv_1, \vv_2, \vv_3\}$.

\end{proof}

\vspace{+2ex}
\noindent
\fbox{\parbox{\textwidth}{
\paragraph{Transition symmetric}
\begin{equation}
\begin{aligned}
\text{Vector-Frame}(\vx_i + \delta \vx, \vx_j + \delta \vx)
& = \textbf{Gram-Schmidt}\{ \vx_i -\vx_j, \vx_i - \vx_k, (\vx_i -\vx_j) \times (\vx_i - \vx_k)\}.
\end{aligned}
\end{equation}
}}

\begin{proof}
Because the basis is based on the difference of coordinates, it is straightforward to observe that $\textbf{Gram-Schmidt}\{\vv_1+\vt, \vv_2+\vt, \vv_3+\vt\} = \textbf{Gram-Schmidt}\{\vv_1, \vv_2, \vv_3\}$. So the frame operation is transition equivariant. We also want to highlight that for all the other vector frame bases introduced in the main article, we remove the mass center for each molecular system, thus, we can guarantee the transition equivariance property.
\end{proof}

\vspace{+2ex}
\noindent
\fbox{\parbox{\textwidth}{
\paragraph{Reflection antisymmetric}
\begin{equation}
\begin{aligned}
\text{Vector-Frame}(\vx_i, \vx_j)
& \ne \text{Vector-Frame}(-\vx_i, -\vx_j).
\end{aligned}
\end{equation}
}}

\begin{proof}
From intuition, this makes sense because the cross-product is anti-symmetric.

A simple counter-example is the original geometry $R$ and the reflected geometry by the original point $-R$. Thus the two bases before and after the reflection group is the following:
\begin{align}
& \textbf{Gram-Schmidt}\{ \vx_i -\vx_j, \vx_i - \vx_k, (\vx_i -\vx_j) \times (\vx_i - \vx_k)\}\\
& \textbf{Gram-Schmidt}\{ -\vx_i +\vx_j, -\vx_i + \vx_k, (\vx_i -\vx_j) \times (\vx_i - \vx_k)\}.
\end{align}
The bases between ${\vv_1, \vv_2, \vv_3}$ and $\{-\vv_1, -\vv_2, \vv_3\}\}$ are different, thus such frame construction is reflection anti-symmetric.

\end{proof}

If you are able to get the above derivations, then you can tell that this can be trivially generalized to arbitrary vector frames as long as the three bases are non-coplanar.

\subsection{Scalarization}
Once we have the three vectors as the vector frame basis, the next step is modeling. Scalarization refers to the function in which we map the vectors to the frames or bases we construct. Suppose the frame is $\mathcal{F} = (\ve_1, \ve_2, \ve_3)$, then for a vector (tensor) $\vh$, the corresponding scalarization is:
\begin{equation}
\vh \odot \mathcal{F} = (\vh \odot \ve_1, \vh \odot \ve_2, \vh \odot \ve_3) = (\vh_1, \vh_2, \vh_3).
\end{equation}

\clearpage
\newpage
\section{MISATO Dataset Specifications}

In this section, we provide more details on the MISATO dataset~\cite{siebenmorgen2024misato}. Note that for small molecule ligands, we ignore the Hydrogen atoms.

\begin{figure}[h]
\centering
\begin{subfigure}[b]{0.32\textwidth}
    \centering
    \includegraphics[width=\textwidth]{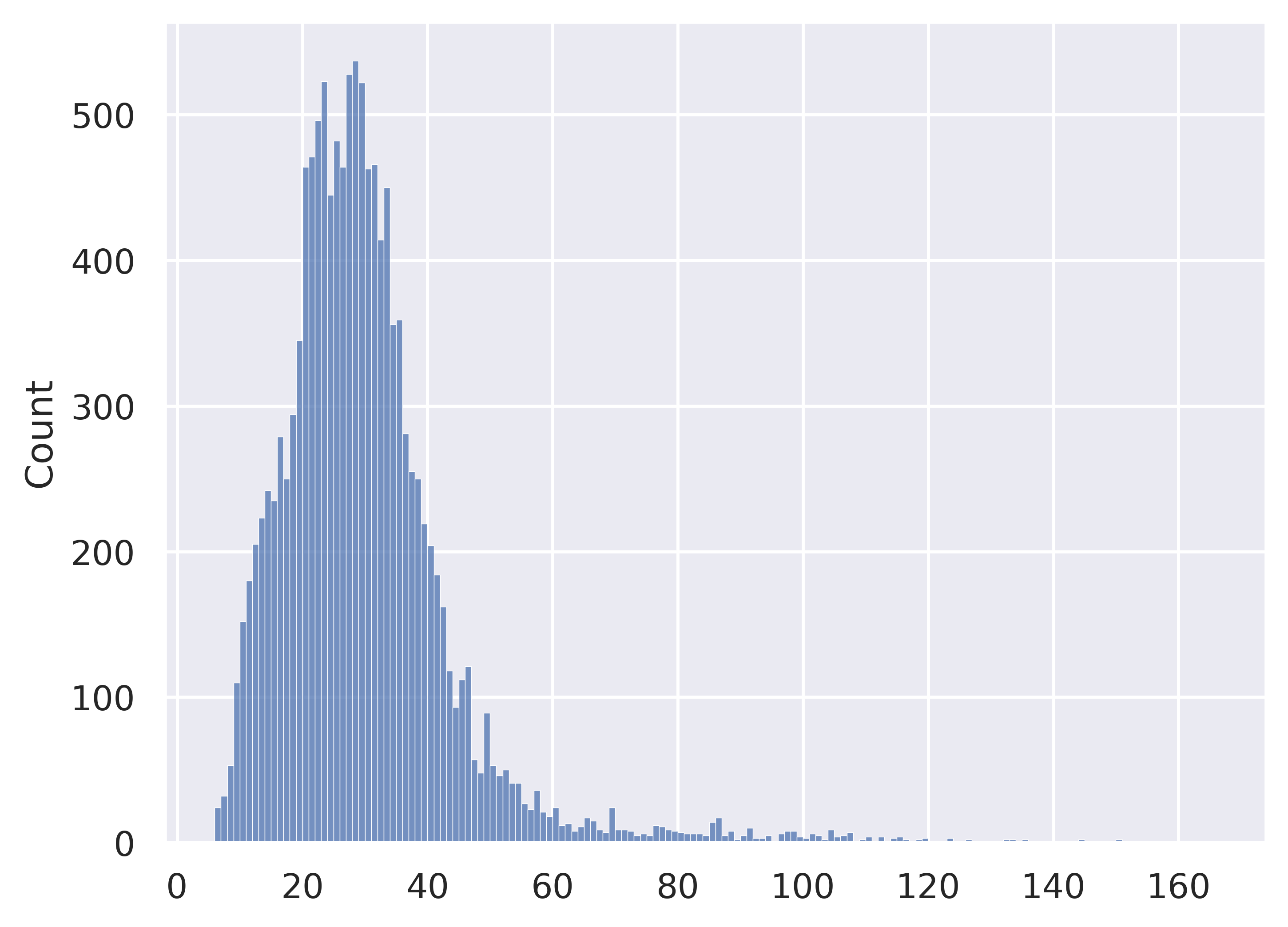}
    \caption{\small Training data.}
\end{subfigure}
\hfill
\begin{subfigure}[b]{0.32\textwidth}
    \centering
    \includegraphics[width=\textwidth]{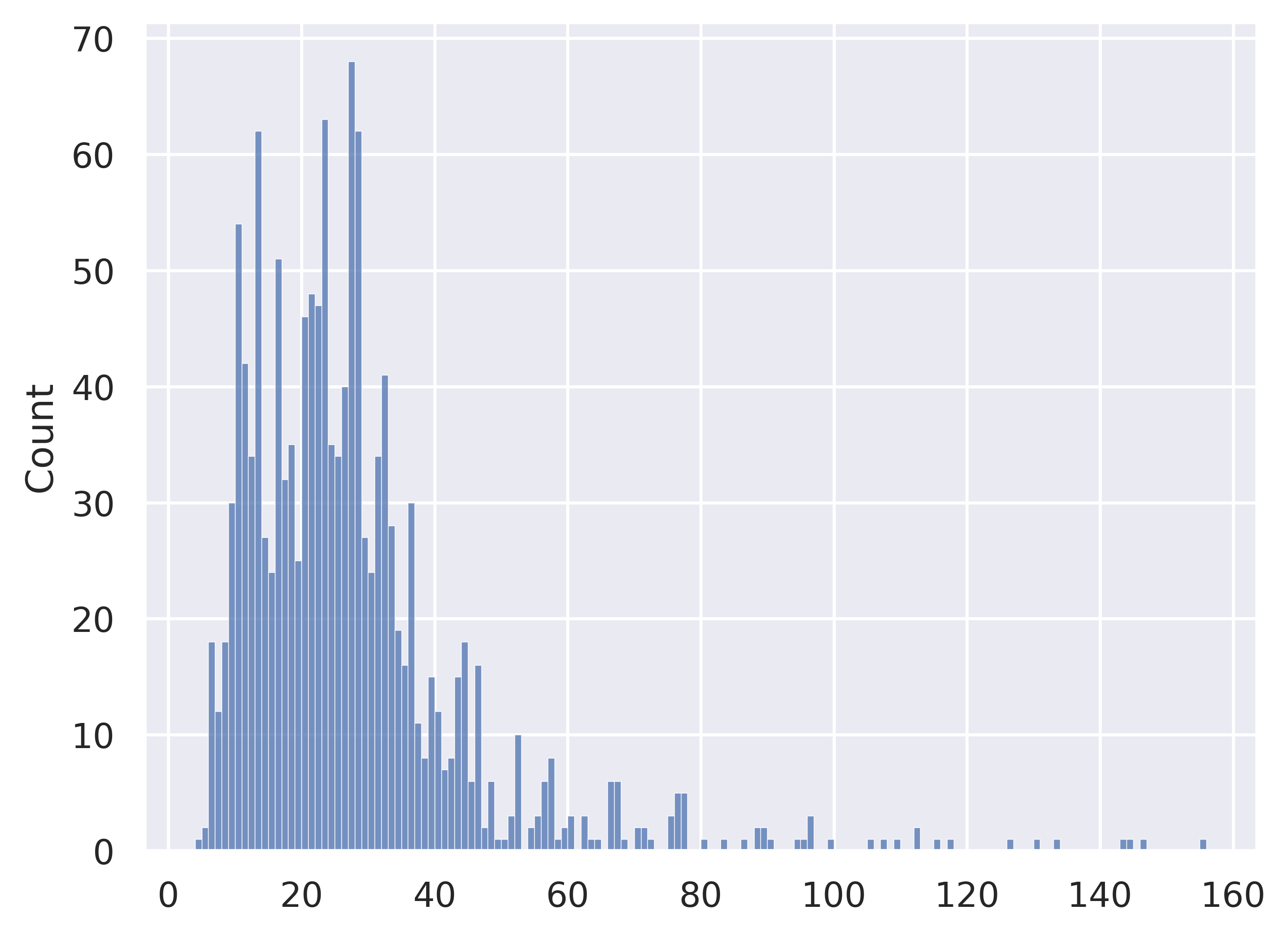}
    \caption{\small Validation data.}
\end{subfigure}
\hfill
\begin{subfigure}[b]{0.32\textwidth}
    \centering
    \includegraphics[width=\textwidth]{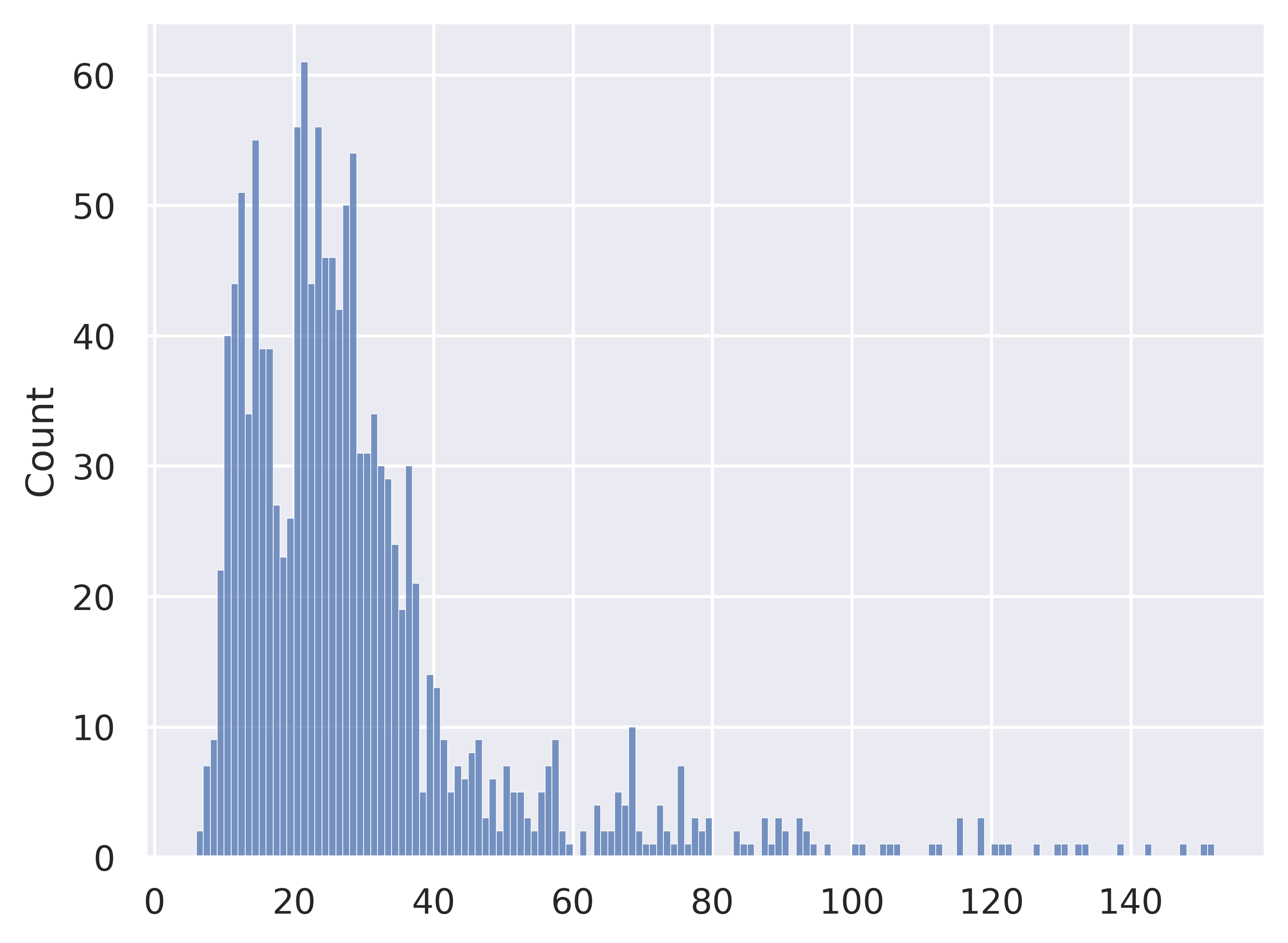}
    \caption{\small Test data.}
\end{subfigure}
\vspace{-2ex}
\caption{\small Distribution on \# atoms in small molecule ligands for all protein-ligand complex.}
\label{fig:atom_number_distribution}
\end{figure}

\begin{figure}[h]
\centering
\begin{subfigure}[b]{0.32\textwidth}
    \centering
    \includegraphics[width=\textwidth]{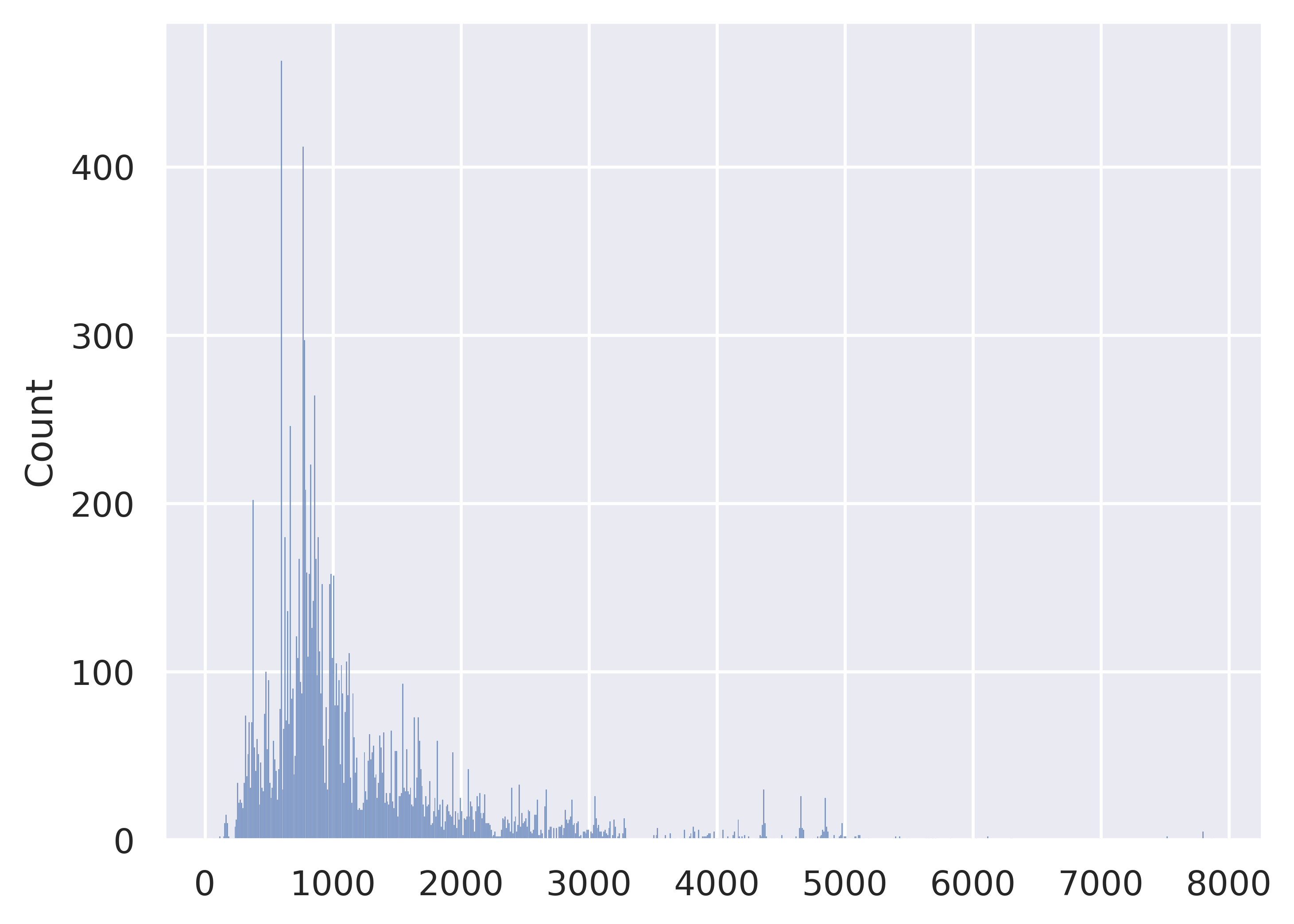}
    \caption{\small Training data.}
\end{subfigure}
\hfill
\begin{subfigure}[b]{0.32\textwidth}
    \centering
    \includegraphics[width=\textwidth]{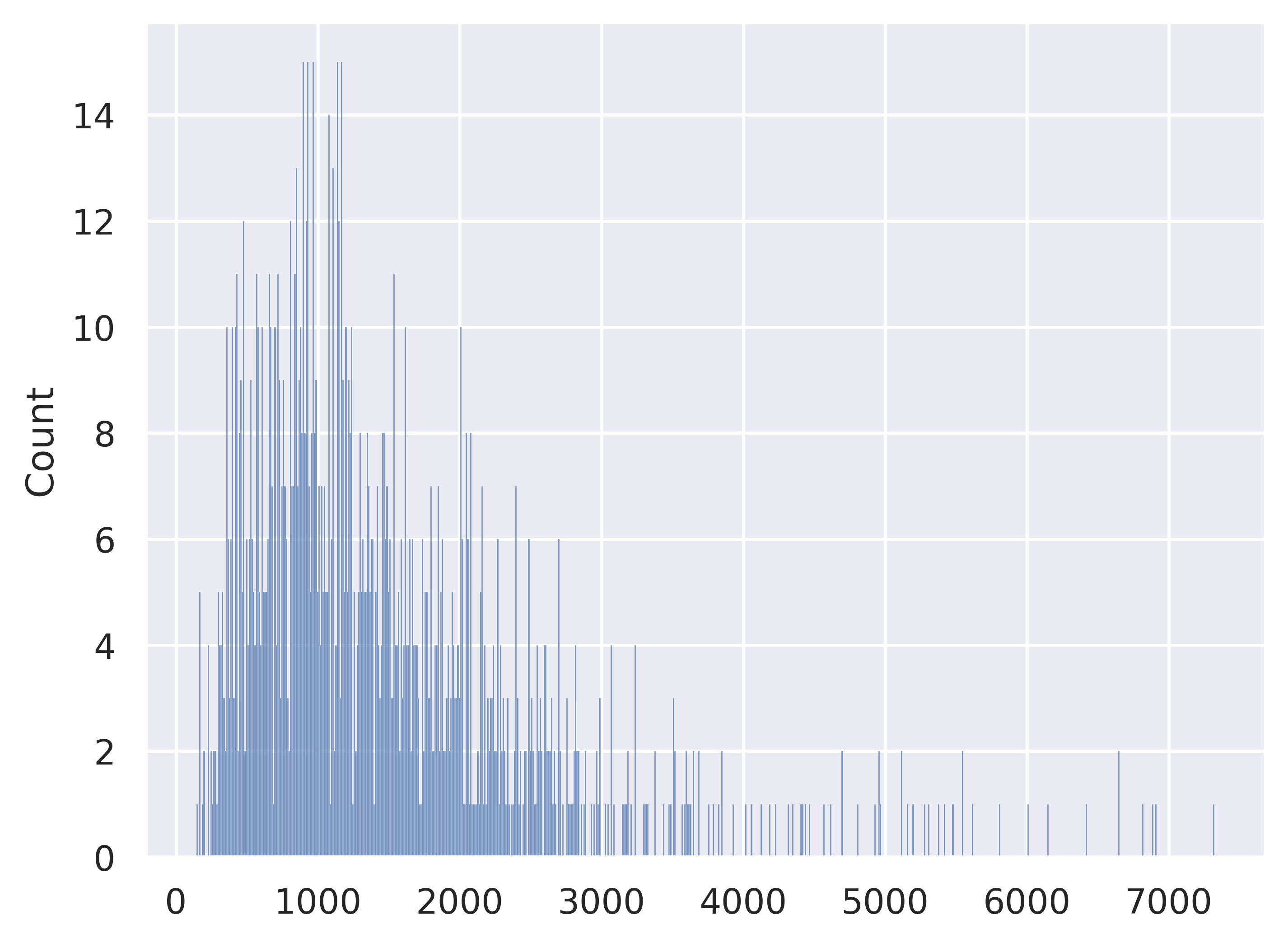}
    \caption{\small Validation data.}
\end{subfigure}
\hfill
\begin{subfigure}[b]{0.32\textwidth}
    \centering
    \includegraphics[width=\textwidth]{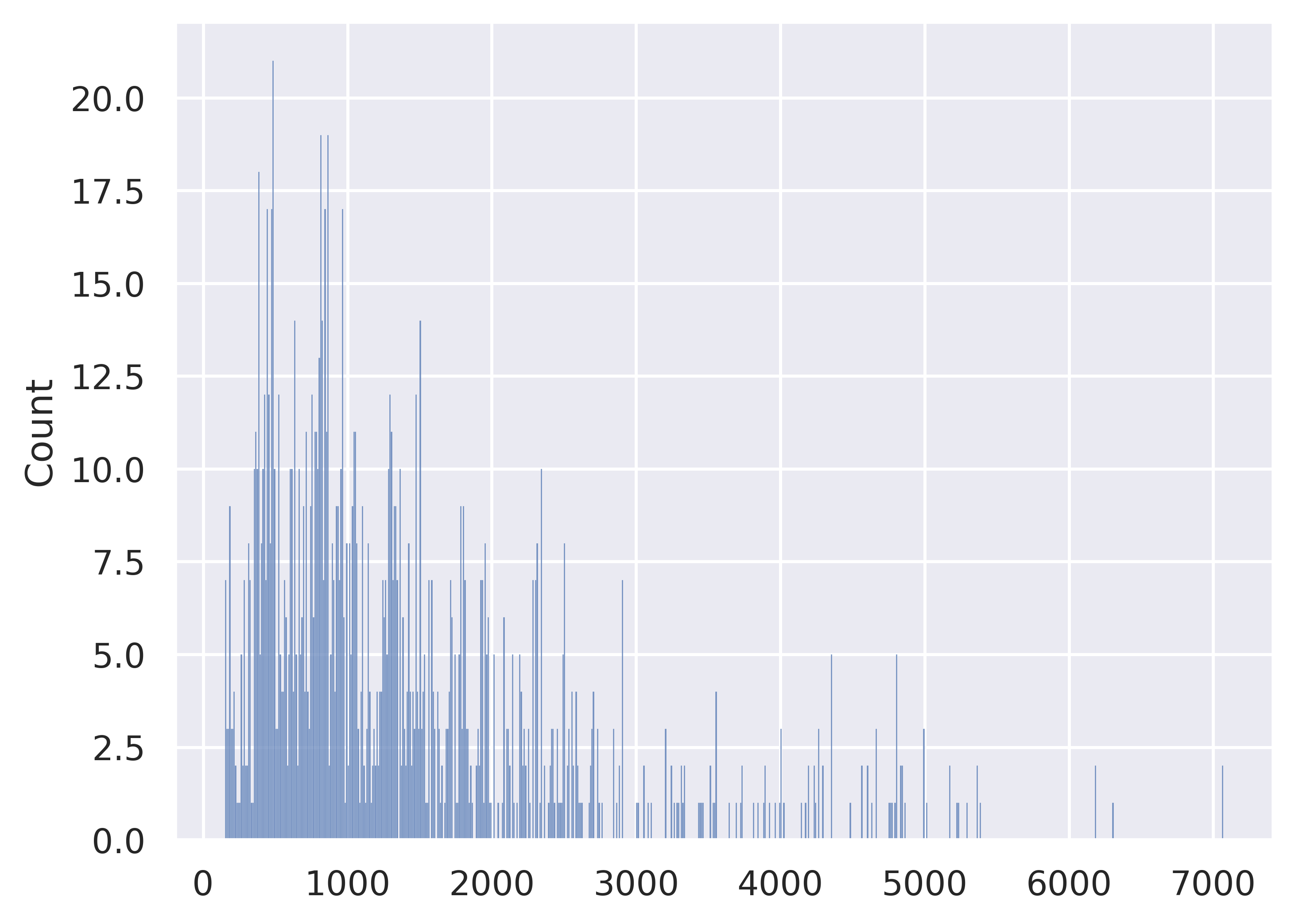}
    \caption{\small Test data.}
\end{subfigure}
\vspace{-2ex}
\caption{\small Distribution on \# residues in proteins for all protein-ligand complex.}
\label{fig:atom_residues_distribution}
\end{figure}

\clearpage
\newpage
\section{Details of \MDModel{}}

\subsection{Model Architecture and Hyperparameters}

\begin{figure}[b]
\centering
\centering
\includegraphics[width=1.\textwidth]{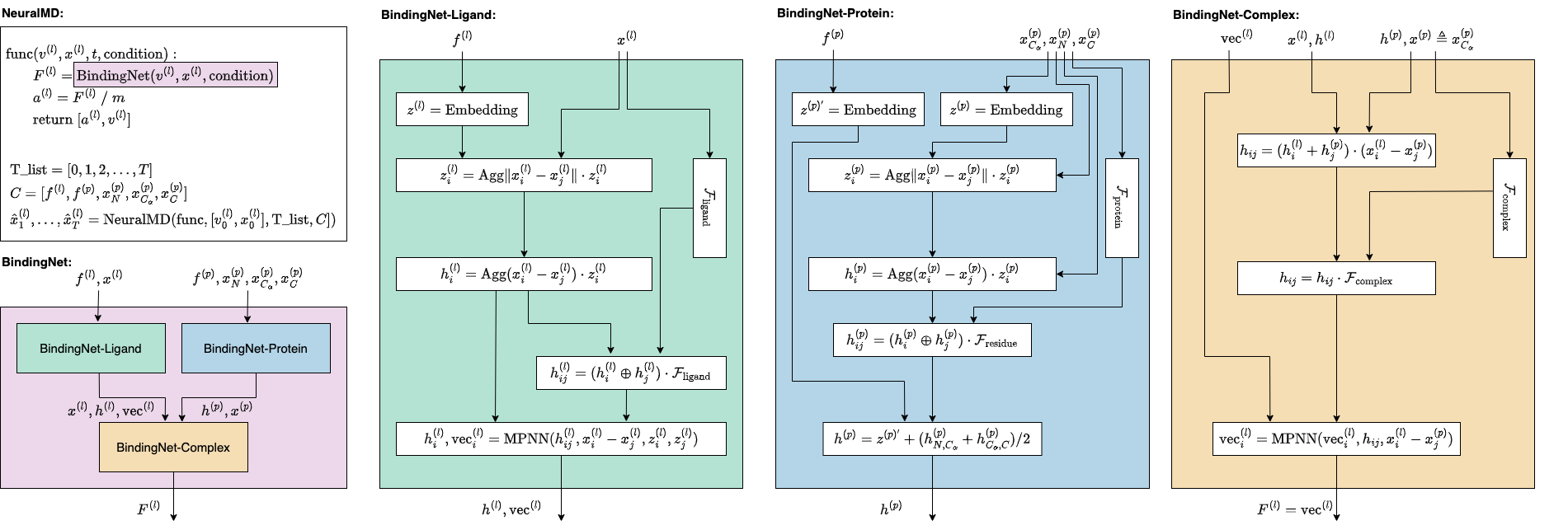}
\vspace{-4ex}
\caption{\small Detailed pipeline of \MDModel{} ODE. In the three key modules of \BindingNet{}, there are three vertical boxes, corresponding to three granularities of vector frames.
}
\label{fig:BindingNet_pipeline}
\end{figure}

In this section, we provide more details on the model architecture in~\Cref{fig:BindingNet_pipeline}, and hyperparameter details in~\Cref{tab:hyperparameter}.

First, we explain each of the three modules in detail and list the dimensions of each variable to make it easier for readers to understand. Suppose the representation dimension is $d$.

\textbf{BindingNet-Ligand: }
\begin{itemize}[noitemsep,topsep=0pt]
    \item $\vz^{(l)} = \text{Embedding}(f^{(l)}) \in \mathbb{R}^{N_{\text{atom}} \cdot d}$ is atom type embedding.
    \item Then for each atom type embedding $\vz^{(l)}$, we add a normalization by multiplying it with the RBF of distance among the neighborhoods, and the resulting atom type embedding stays the same dimension $\vz^{(l)} \in \mathbb{R}^{N_{\text{atom}} \times d}$.
    \item $\{\vh^{(l)}_i = \text{Agg}_j (\vx_i^{(l)} - \vx_j^{(l)}) \cdot \vz_i^{(l)} \} \in \mathbb{R}^{N_{\text{atom}} \cdot d \cdot 3}$ is the equivariant representation of each atom.
    \item $\{\vh^{(l)}_{ij} \} \in \mathbb{R}^{N_{\text{edge}} \cdot 2d \cdot 3}$ is the invariant representation after scalarization. Then we will take a simple sum-pooling, followed by an MLP to get the invariant representation $\vh^{(l)}_{ij} \in \mathbb{R}^{N_{\text{edge}} \cdot d}$.
    \item Finally, we will repeat $L$ layers of MPNN:
    \begin{equation}
    \begin{aligned}
        \text{vec}^{(l)}_i &= \text{vec}^{(l)}_i + \text{Agg}_{j} \big( \text{vec}^{(l)}_i \cdot \text{MLP}(\vh_{ij}) + (\vx^{(l)}_i - \vx^{(p)}_j) \cdot \text{MLP}(\vh_{ij}) \big), \;\;\; //\{\text{vec}^{(l)}_i\} \in \mathbb{R}^{N_{\text{atom}} \cdot 3}\\
        \vh^{(l)}_i &= \vh^{(l)}_i + \text{Agg}_{j} \big(\text{MLP}(\vh_{ij}) \big). \;\;\; //\{\vh^{(l)}_i\} \in\mathbb{R}^{N_{\text{atom}} \cdot d}
    \end{aligned}
    \end{equation}
\end{itemize}

\textbf{BindingNet-Protein:}
\begin{itemize}[noitemsep,topsep=0pt]
    \item $\vz^{(p)} \in \mathbb{R}^{N_{\text{backbone-atom}} \cdot d}$ is the \text{backbone-atom} type representation by aggregating the neighbors without the cutoff $c$.
    \item $\tilde \vz^{(p)} \in \mathbb{R}^{N_{\text{backbone-atom}} \cdot d}$ is the \text{backbone-atom} type representation.
    \item $\{\vh_i^{(p)}\} \in \mathbb{R}^{N_{\text{backbone-atom}} \cdot d \cdot 3}$ is the \text{backbone-atom} equivariant representation.
    \item $\{\vh_{ij}^{(p)}\} \in \mathbb{R}^{N_{\text{edge}} \cdot 2d \cdot 3}$ is the invariant representation after scalarization. Then we take a simple sum-pooling, followed by an MLP to get the invariant representation $\{ \vh^{(l)}_{ij} \} \in \mathbb{R}^{N_{\text{edge}} \cdot d}$.
    \item Finally, we get the residue-level representation as $\vh^{(p)} = \tilde \vz^{(p)} + (\vh^{(p)}_{N,C_\alpha} + \vh^{(p)}_{C_\alpha, C})/2 \in \mathbb{R}^{N_{\text{residue}} \cdot d}$.
\end{itemize}

\textbf{BindingNet-Complex:}
\begin{itemize}[noitemsep,topsep=0pt]
    \item $\{\vh_{ij}\} \in \mathbb{R}^{N_{\text{edge}} \cdot d \cdot 3}$ is the equivariant interaction/edge representation.
    \item $\{\vh_{ij} = \vh_{ij} \cdot \mathcal{F}_{\text{complex}}\} \mathbb{R}^{N_{\text{edge}} \cdot d \cdot 3}$ is the scalarization. Then we take a simple sum-pooling, followed by an MLP to get the invariant representation $\{\vh^{(l)}_{ij}\} \in \mathbb{R}^{N_{\text{edge}} \cdot d}$.
    \item The final output is obtained by $L$ MPNN layers as:
    \begin{equation}
    \begin{aligned}
    \text{vec}^{(pl)}_{ij} &= \text{vec}^{(l)}_i \cdot \text{MLP}(h_{ij}) + (\vx^{(l)}_i - \vx^{(p)}_j) \cdot \text{MLP}(h_{ij}), \;\;\; //\{\text{vec}^{(pl)}_{ij}\} \in \mathbb{R}^{N_{\text{edge}} \cdot 3}\\
    F^{(l)}_i &= \text{vec}^{(l)}_i + \text{Agg}_{j \in \mathcal{N}(i)} \text{vec}^{(pl)}_{ij}. \;\;\; //\{F^{(l)}_i\} \in \mathbb{R}^{N_{\text{atom}} \cdot 3}
    \end{aligned}
    \end{equation}
\end{itemize}

\begin{table}[H]
\centering
\setlength{\tabcolsep}{5pt}
\fontsize{9}{9}\selectfont
\caption{
\small
Hyperparameter specifications for \MDModel{}.
}
\label{tab:hyperparameter}
\vspace{-2ex}
\begin{adjustbox}{max width=\textwidth}
\begin{tabular}{l l l}
\toprule
Hyperparameter & Value\\
\midrule
\# layers & $\{5\}$\\
cutoff $c$ & $\{5\}$\\
velocity initial mapping function & \{True, False\}\\
velocity refinement coefficient $\alpha$ & \{0, 0.01, 0.001\}\\
step size (integration) & \{0.25, 0.5, 1\}\\
learning rate & \{1e-3, 1e-4\} \\
optimizer & \{SGD, Adam \}\\
\bottomrule
\end{tabular}
\end{adjustbox}
\end{table}

\subsection{Overdamped, Underdamped, and Data-driven Langevin Dynamics}
In the literature on using Langevin dynamics for MD simulation, there are two main categories: \textbf{overdamped Langevin dynamics} and \textbf{underdamped Langevin dynamics}. In this section, we discuss these methods and their applicable settings.

\paragraph{Langevin dynamics} or damped Langevin dynamics is defined as
\begin{equation} \label{eq:general_Langevin_dynamics}
\begin{aligned}
    m \va = -\nabla U(\vx) - \gamma m \vv + \sqrt{2m \gamma k_BT} R(t), 
\end{aligned}
\end{equation}
where $\gamma$ is the damping constant or collision frequency, $T$ is the temperature, $k_B$ is the Boltzmann's constant, and $R(t)$ is a delta-correlated stationary Gaussian process with zero-mean.

\paragraph{Overdamped Langevin dynamics (Brownian dynamics)}
This is the friction-dominated regime, where the inertia term ($m \va$) is negligible compared to the friction term ($-\gamma m \vv$). The equation for overdamped Langevin dynamics is:
\begin{equation}
\begin{aligned}
-\nabla U(\vx) - \gamma m \vv + \sqrt{2m \gamma k_BT} R(t) = 0.
\end{aligned}
\end{equation}
Thus, the trajectories are given by:
\begin{equation}
\begin{aligned}
\vx_{t+1} - \vx_{t} & = - \frac{1}{\gamma m} \nabla U(x) + \frac{\sqrt{2m \gamma k_BT}}{\gamma m} R(t)\\
& = -\frac{D}{k_B T} \nabla U(X) + \sqrt{2D} R(t),
\end{aligned}
\end{equation}
where $D = k_B T / \gamma$.
This has been widely used in deep generative models like Denoising Diffusion Probabilistic Model (DDPM)~\cite{ho2020denoising}. Specifically, it is employed as the Monte Carlo sampling step.

For application, this is suitable for large molecular systems like protein folding in solution because the motion is slow and dominated by viscous drag. However, such a dynamic is not suitable for simulating small particles like small molecules.

\paragraph{Underdamped Langevin dynamics}
This is the inertia-dominated regime, where the inertia term ($m \va$) is comparable to or larger than the friction term ($- \gamma m \vv$). It is applicable for describing a particle moving in a low-viscosity medium or when the friction coefficient $\gamma$ is small, allowing the particle to exhibit significant inertial motion.

For application, underdamped Langevin dynamics is ideal for systems where inertia plays an important role, such as in small molecules or systems with oscillations or high-frequency dynamics. This regime is typically used in molecular dynamics (MD) simulations where inertia is significant, such as simulating vibrational modes of molecules or dynamics in a gas phase.\looseness=-1

\paragraph{Data-driven Langevin dynamics in \MDModel{}}
The Langevin dynamics modeled in \MDModel{} make no prior assumptions about overdamped or underdamped behavior. Instead, we learn a data-driven approximation of \Cref{eq:general_Langevin_dynamics}:
\begin{equation}
    m\va = \text{\BindingNet{}}(f^{(l)}, \vx^{(l)}, f^{(p)}, \vx^{(p)}_N, \vx^{(p)}_{C_\alpha}, \vx^{(p)}_C) + \text{\BindingNet{}-Ligand}(f^{(l)}, \vx^{(l)}) \cdot \epsilon,
\end{equation}
where we are using the reparameterization trick, and $\epsilon$ is sampled from a standard Gaussian.

\paragraph{Summary}
To summarize, if we treat the MD simulation of small particles as a density estimation task, there are several solutions available, including DDPM-like DenoisingLD and \MDModel{}. However, if we consider these models as learning or simulating dynamics from a physics perspective, we must be cautious. Current DDPM methods like DenoisingLD rely on overdamped Langevin dynamics, which are not well-suited for simulating small molecule dynamics. In this context, \MDModel{} is a more accurate option.

\clearpage
\newpage
\section{Empirical Results}

\subsection{Standard Deviation for Multi-trajectory Semi-flexible Binding Prediction}
We are running all the experiments using three random seeds, 0, 42, and 123. In the tables of the main article, we are only reporting the mean due to the space limitation. Thus here, we would also like to report the results with standard deviation.
We show the reconstruction and validity results on three multi-trajectory binding dynamics: MISATO-100 in \Cref{tab:complete_result_multi_trajectory_100}, MISATO-1000 in \Cref{tab:complete_result_multi_trajectory_1000}, and MISATO-All in \Cref{tab:complete_result_multi_trajectory_All}.

\begin{table}[h]
\setlength{\tabcolsep}{15pt}
\fontsize{9}{9}\selectfont
\centering
\caption{
\small
Results of multi-trajectory binding dynamics predictions on MISATO-100. Four evaluation metrics are considered: MAE (\AA, $\downarrow$), MSE ($\downarrow$), Matching($\downarrow$), Matching ($\downarrow$), and Stability (\%, $\uparrow$).
}
\label{tab:complete_result_multi_trajectory_100}
\vspace{-2ex}
\begin{adjustbox}{max width=\textwidth}
\begin{tabular}{l rrrr rrrr rrrr}
\toprule
 & \multicolumn{2}{c}{Reconstruction} & \multicolumn{2}{c}{Validity}
 \\
\cmidrule(lr){2-3} \cmidrule(lr){4-5}
 & MAE & MSE & Matching & Stability
 \\
\midrule
VerletMD & 85.286 $\pm$ 0.035 & 54.996 $\pm$ 0.17 & 46.753 $\pm$ 2.05 & 10.051 $\pm$ 1.39\\
GNN-MD & 5.964 $\pm$ 0.05 & 3.938 $\pm$ 0.02 & 0.671 $\pm$ 0.02 & 70.546 $\pm$ 1.81\\
DenoisingLD & 8.251 $\pm$ 0.02 & 5.541 $\pm$ 0.01 & 1.744 $\pm$ 0.04 & 29.545 $\pm$ 0.68\\
\midrule
NeuralMD ODE (ours) & 5.867 $\pm$ 0.00 & 3.870 $\pm$ 0.00 & 0.539 $\pm$ 0.02 & 79.553 $\pm$ 2.17\\
NeuralMD SDE (ours) & 5.868 $\pm$ 0.00 & 3.871 $\pm$ 0.00 & 0.533 $\pm$ 0.02 & 80.229 $\pm$ 1.23\\
\bottomrule
\end{tabular}
\end{adjustbox}
\end{table}

\begin{table}[h]
\setlength{\tabcolsep}{15pt}
\fontsize{9}{9}\selectfont
\centering
\caption{
\small
Results of multi-trajectory binding dynamics predictions on MISATO-1000. Four evaluation metrics are considered: MAE (\AA, $\downarrow$), MSE ($\downarrow$), Matching($\downarrow$), Matching ($\downarrow$), and Stability (\%, $\uparrow$).
}
\label{tab:complete_result_multi_trajectory_1000}
\vspace{-2ex}
\begin{adjustbox}{max width=\textwidth}
\begin{tabular}{l rrrr rrrr rrrr}
\toprule
 & \multicolumn{2}{c}{Reconstruction} & \multicolumn{2}{c}{Validity}
 \\
\cmidrule(lr){2-3} \cmidrule(lr){4-5}
 & MAE & MSE & Matching & Stability
 \\
\midrule
VerletMD & 104.537 $\pm$ 0.07 & 68.942 $\pm$ 0.04 & 48.899 $\pm$ 0.19 & 10.574 $\pm$ 0.59\\
GNN-MD & 7.524 $\pm$ 0.01 & 4.915 $\pm$ 0.01 & 0.670 $\pm$ 0.05 & 68.310 $\pm$ 4.08\\
DenoisingLD & 9.251 $\pm$ 0.15 & 6.074 $\pm$ 0.10 & 1.362 $\pm$ 0.06 & 37.289 $\pm$ 1.54\\
\midrule
NeuralMD ODE (ours) & 7.459 $\pm$ 0.00 & 4.867 $\pm$ 0.00 & 0.612 $\pm$ 0.02 & 70.362 $\pm$ 1.99\\
NeuralMD SDE (ours) & 7.476 $\pm$ 0.00 & 4.876 $\pm$ 0.00 & 0.457 $\pm$ 0.00 & 83.960 $\pm$ 0.00\\
\bottomrule
\end{tabular}
\end{adjustbox}
\end{table}

\begin{table}[h]
\setlength{\tabcolsep}{15pt}
\fontsize{9}{9}\selectfont
\centering
\caption{
\small
Results of multi-trajectory binding dynamics predictions on MISATO-All. Four evaluation metrics are considered: MAE (\AA, $\downarrow$), MSE ($\downarrow$), Matching($\downarrow$), Matching ($\downarrow$), and Stability (\%, $\uparrow$).
}
\label{tab:complete_result_multi_trajectory_All}
\vspace{-2ex}
\begin{adjustbox}{max width=\textwidth}
\begin{tabular}{l rrrr rrrr rrrr}
\toprule
 & \multicolumn{2}{c}{Reconstruction} & \multicolumn{2}{c}{Validity}
 \\
\cmidrule(lr){2-3} \cmidrule(lr){4-5}
 & MAE & MSE & Matching & Stability
 \\
\midrule
VerletMD & 97.213 $\pm$ 0.29 & 64.405 $\pm$ 0.19 & 50.857 $\pm$ 0.67 & 11.888 $\pm$ 1.49\\
GNN-MD & 7.637 $\pm$ 0.01 & 5.048 $\pm$ 0.00 & 0.675 $\pm$ 0.02 & 69.244 $\pm$ 1.65\\
DenoisingLD & 8.149 $\pm$ 0.79 & 5.387 $\pm$ 0.52 & 0.764 $\pm$ 0.38 & 68.315 $\pm$ 20.01\\
\midrule
NeuralMD ODE (ours) & 7.513 $\pm$ 0.01 & 4.961 $\pm$ 0.00 & 0.491 $\pm$ 0.02 & 81.991 $\pm$ 1.80\\
NeuralMD SDE (ours) & 7.517 $\pm$ 0.00 & 4.963 $\pm$ 0.00 & 0.474 $\pm$ 0.00 & 83.264 $\pm$ 0.00\\
\bottomrule
\end{tabular}
\end{adjustbox}
\end{table}

\clearpage
\newpage

\subsection{Standard Deviation for Single-trajectory Semi-flexible Binding Prediction}

We illustrate the reconstruction and validity results with standard deviation on ten single-trajectory binding dynamics in \Cref{tab:complete_result_single_trajectory}.

\begin{table}[h]
\centering
\setlength{\tabcolsep}{5pt}
\caption{
\small
Results on ten single-trajectory binding dynamics prediction. 
Results with optimal training loss are reported.
Four evaluation metrics are considered: MAE (\AA, $\downarrow$), MSE ($\downarrow$), Matching ($\downarrow$), and Stability (\%, $\downarrow$).
}
\label{tab:complete_result_single_trajectory}
\vspace{-2ex}
\centering
\begin{adjustbox}{max width=0.8\textwidth}
\begin{tabular}{l l rrr rr}
\toprule
 &  & VerletMD & GNN MD & Denoising LD & NeuralMD (ODE) & NeuralMD (SDE)\\
\midrule
\multirow{6}{*}{5WIJ} & MAE & 14.629 $\pm$ 0.04 & 2.280 $\pm$ 0.05 & 2.501 $\pm$ 0.01 & 2.252 $\pm$ 0.16 & 2.260 $\pm$ 0.15\\
 & MSE & 10.221 $\pm$ 0.03 & 1.521 $\pm$ 0.03 & 1.644 $\pm$ 0.00 & 1.514 $\pm$ 0.13 & 1.514 $\pm$ 0.10\\
 & Matching & 5.459 $\pm$ 0.10 & 0.803 $\pm$ 0.09 & 0.815 $\pm$ 0.07 & 0.464 $\pm$ 0.09 & 0.615 $\pm$ 0.18\\
 & Stability & 24.360 $\pm$ 0.52 & 54.475 $\pm$ 2.45 & 52.418 $\pm$ 4.65 & 82.046 $\pm$ 6.26 & 67.464 $\pm$ 14.15\\
\midrule
\multirow{6}{*}{4ZX0} & MAE & 21.278 $\pm$ 0.02 & 2.370 $\pm$ 0.09 & 3.138 $\pm$ 0.06 & 1.878 $\pm$ 0.00 & 2.158 $\pm$ 0.17\\
 & MSE & 14.357 $\pm$ 0.01 & 1.599 $\pm$ 0.05 & 2.045 $\pm$ 0.03 & 1.263 $\pm$ 0.00 & 1.455 $\pm$ 0.14\\
 & Matching & 7.971 $\pm$ 0.05 & 0.555 $\pm$ 0.02 & 1.072 $\pm$ 0.03 & 0.428 $\pm$ 0.00 & 0.696 $\pm$ 0.09\\
 & Stability & 19.168 $\pm$ 0.35 & 68.613 $\pm$ 1.14 & 44.228 $\pm$ 0.78 & 81.401 $\pm$ 0.78 & 59.109 $\pm$ 7.87\\
\midrule
\multirow{6}{*}{3EOV} & MAE & 27.960 $\pm$ 0.06 & 3.512 $\pm$ 0.02 & 4.055 $\pm$ 0.01 & 3.858 $\pm$ 0.31 & 3.395 $\pm$ 0.22\\
 & MSE & 18.821 $\pm$ 0.04 & 2.413 $\pm$ 0.02 & 2.787 $\pm$ 0.01 & 2.651 $\pm$ 0.20 & 2.309 $\pm$ 0.11\\
 & Matching & 13.588 $\pm$ 0.12 & 1.216 $\pm$ 0.05 & 1.209 $\pm$ 0.01 & 1.062 $\pm$ 0.06 & 0.962 $\pm$ 0.03\\
 & Stability & 13.067 $\pm$ 0.29 & 40.984 $\pm$ 2.58 & 41.469 $\pm$ 0.81 & 47.328 $\pm$ 4.95 & 50.108 $\pm$ 3.48\\
\midrule
\multirow{6}{*}{4K6W} & MAE & 15.428 $\pm$ 0.01 & 3.695 $\pm$ 0.04 & 3.942 $\pm$ 0.01 & 3.656 $\pm$ 0.12 & 3.765 $\pm$ 0.20\\
 & MSE & 10.357 $\pm$ 0.01 & 2.402 $\pm$ 0.03 & 2.635 $\pm$ 0.00 & 2.400 $\pm$ 0.08 & 2.501 $\pm$ 0.17\\
 & Matching & 7.505 $\pm$ 0.06 & 1.038 $\pm$ 0.07 & 0.839 $\pm$ 0.01 & 0.928 $\pm$ 0.19 & 1.076 $\pm$ 0.49\\
 & Stability & 15.441 $\pm$ 0.13 & 42.480 $\pm$ 2.74 & 53.820 $\pm$ 0.57 & 49.438 $\pm$ 8.66 & 49.700 $\pm$ 12.15\\
\midrule
\multirow{6}{*}{1KTI} & MAE & 18.157 $\pm$ 0.04 & 6.641 $\pm$ 0.05 & 7.051 $\pm$ 0.02 & 6.675 $\pm$ 0.04 & 6.646 $\pm$ 0.01\\
 & MSE & 12.723 $\pm$ 0.03 & 4.173 $\pm$ 0.04 & 4.369 $\pm$ 0.01 & 4.176 $\pm$ 0.05 & 4.141 $\pm$ 0.01\\
 & Matching & 7.467 $\pm$ 0.03 & 0.386 $\pm$ 0.03 & 0.268 $\pm$ 0.02 & 0.337 $\pm$ 0.17 & 0.167 $\pm$ 0.02\\
 & Stability & 19.352 $\pm$ 0.61 & 81.831 $\pm$ 2.03 & 91.986 $\pm$ 2.36 & 86.430 $\pm$ 11.59 & 98.508 $\pm$ 0.57\\
\midrule
\multirow{6}{*}{1XP6} & MAE & 13.753 $\pm$ 0.01 & 2.378 $\pm$ 0.06 & 2.218 $\pm$ 0.03 & 1.924 $\pm$ 0.05 & 2.061 $\pm$ 0.06\\
 & MSE & 9.587 $\pm$ 0.01 & 1.561 $\pm$ 0.03 & 1.472 $\pm$ 0.02 & 1.280 $\pm$ 0.04 & 1.356 $\pm$ 0.05\\
 & Matching & 4.672 $\pm$ 0.08 & 0.966 $\pm$ 0.08 & 0.676 $\pm$ 0.01 & 0.537 $\pm$ 0.04 & 0.615 $\pm$ 0.11\\
 & Stability & 28.129 $\pm$ 1.17 & 49.239 $\pm$ 2.78 & 64.951 $\pm$ 0.33 & 75.533 $\pm$ 3.82 & 69.423 $\pm$ 6.48\\
\midrule
\multirow{6}{*}{4YUR} & MAE & 16.764 $\pm$ 0.03 & 7.031 $\pm$ 0.07 & 7.128 $\pm$ 0.01 & 6.957 $\pm$ 0.07 & 7.038 $\pm$ 0.19\\
 & MSE & 11.069 $\pm$ 0.00 & 4.641 $\pm$ 0.07 & 4.807 $\pm$ 0.00 & 4.597 $\pm$ 0.03 & 4.679 $\pm$ 0.11\\
 & Matching & 9.555 $\pm$ 0.04 & 0.920 $\pm$ 0.02 & 0.834 $\pm$ 0.01 & 0.584 $\pm$ 0.07 & 0.749 $\pm$ 0.27\\
 & Stability & 16.542 $\pm$ 0.42 & 47.555 $\pm$ 1.06 & 49.676 $\pm$ 0.64 & 69.775 $\pm$ 6.88 & 60.344 $\pm$ 16.47\\
\midrule
\multirow{6}{*}{4G3E} & MAE & 5.111 $\pm$ 0.02 & 2.709 $\pm$ 0.08 & 3.588 $\pm$ 0.11 & 2.191 $\pm$ 0.02 & 2.345 $\pm$ 0.26\\
 & MSE & 3.503 $\pm$ 0.01 & 1.785 $\pm$ 0.02 & 2.321 $\pm$ 0.07 & 1.453 $\pm$ 0.01 & 1.536 $\pm$ 0.16\\
 & Matching & 3.388 $\pm$ 0.03 & 0.893 $\pm$ 0.59 & 1.069 $\pm$ 0.09 & 0.505 $\pm$ 0.12 & 0.521 $\pm$ 0.03\\
 & Stability & 31.852 $\pm$ 0.79 & 61.802 $\pm$ 18.90 & 40.823 $\pm$ 4.19 & 71.436 $\pm$ 11.43 & 68.729 $\pm$ 2.78\\
\midrule
\multirow{6}{*}{6B7F} & MAE & 31.934 $\pm$ 0.01 & 4.136 $\pm$ 0.10 & 4.431 $\pm$ 0.01 & 3.921 $\pm$ 0.14 & 3.842 $\pm$ 0.06\\
 & MSE & 22.168 $\pm$ 0.01 & 2.768 $\pm$ 0.08 & 3.047 $\pm$ 0.01 & 2.652 $\pm$ 0.11 & 2.601 $\pm$ 0.03\\
 & Matching & 21.691 $\pm$ 0.02 & 1.194 $\pm$ 0.13 & 0.672 $\pm$ 0.03 & 0.459 $\pm$ 0.14 & 0.741 $\pm$ 0.26\\
 & Stability & 11.050 $\pm$ 0.04 & 39.067 $\pm$ 5.15 & 61.583 $\pm$ 3.10 & 75.692 $\pm$ 11.97 & 57.917 $\pm$ 13.31\\
\midrule
\multirow{6}{*}{3B9S} & MAE & 19.473 $\pm$ 0.04 & 2.578 $\pm$ 0.08 & 2.811 $\pm$ 0.05 & 3.039 $\pm$ 0.52 & 3.132 $\pm$ 0.92\\
 & MSE & 11.696 $\pm$ 0.03 & 1.699 $\pm$ 0.06 & 1.868 $\pm$ 0.04 & 1.999 $\pm$ 0.35 & 2.078 $\pm$ 0.62\\
 & Matching & 0.923 $\pm$ 0.15 & 1.414 $\pm$ 0.27 & 0.472 $\pm$ 0.03 & 0.659 $\pm$ 0.43 & 0.444 $\pm$ 0.15\\
 & Stability & 57.801 $\pm$ 3.84 & 49.306 $\pm$ 11.48 & 71.852 $\pm$ 1.40 & 76.065 $\pm$ 17.78 & 77.801 $\pm$ 11.54\\
\bottomrule
\end{tabular}
\end{adjustbox}
\end{table}

\end{document}